\documentclass[sigconf]{acmart}
\usepackage{enumitem}
\usepackage{bbm}
\usepackage{multirow}
\usepackage{caption}
\usepackage{graphicx}
\usepackage{float} 
\usepackage{subcaption}
\usepackage{colortbl}
\usepackage{balance}
\usepackage{multicol}
\usepackage[linesnumbered, ruled]{algorithm2e}
\SetKwRepeat{Do}{do}{while}%
\usepackage{pifont}
\usepackage{arydshln}
\makeatletter
\DeclareRobustCommand\onedot{\futurelet\@let@token\@onedot}
\def\@onedot{\ifx\@let@token.\else.\null\fi\xspace}

\definecolor{myPink}{RGB}{255, 217, 224}

\def\eg{\emph{e.g}\onedot} 
\def\ie{\emph{i.e}\onedot}

\def\model{\textsc{RAPO}}

\colorlet{tabularColor}{gray!10}

\newcommand{\ours}[1]{\cellcolor{orange!10}#1}

\definecolor{avg}{RGB}{242,242,254}

\definecolor{up}{RGB}{0, 100, 0}
\definecolor{down}{RGB}{139, 0, 0}

\definecolor{l1}{rgb}{0.4, 0.9, 0.75} % 提升 1 以内（较浅绿色）
\definecolor{l2}{rgb}{0.35, 0.8, 0.7} % 提升 1-5
\definecolor{l3}{rgb}{0.3, 0.7, 0.6}  % 提升 5-8
\definecolor{l4}{rgb}{0.25, 0.6, 0.5} % 提升 8-14
\definecolor{l5}{rgb}{0.2, 0.45, 0.4}  % 提升 14 以上（浅绿）
\definecolor{r1}{rgb}{1.0, 0.95, 0.95} % 效果降低
\definecolor{r2}{rgb}{1.0, 0.9, 0.9} % 效果降低

\usepackage{tabularx}
\usepackage[most]{tcolorbox}

\definecolor{think}{HTML}{053CF6}
\definecolor{search}{HTML}{03B6E9}
\definecolor{information}{HTML}{C46F40}
\definecolor{answer}{HTML}{CA033A}
\definecolor{down}{HTML}{B1281C}
\definecolor{up}{HTML}{489638}

\usepackage{minted}

\definecolor{primary_1}{HTML}{4852B1}
\newtcolorbox{templatebox}[1]{
    breakable,
    enhanced,
    colback=white,
    colframe=gray!80!black,
    colbacktitle=gray!80!black,
    coltitle=white,
    fonttitle=\bfseries,
    title=#1,
    arc=3mm,
    boxrule=1pt,
    drop fuzzy shadow={gray!50!white},
    left=5mm,
    right=5mm,
    top=3mm,
    bottom=3mm
}

\makeatother

\AtBeginDocument{%
  \providecommand\BibTeX{{%
    \normalfont B\kern-0.5em{\scshape i\kern-0.25em b}\kern-0.8em\TeX}}}

\setcopyright{acmcopyright}
\copyrightyear{2026}
\acmYear{2026}
\acmDOI{XXXXXXX.XXXXXXX}
\acmConference[KDD'26]{KDD'26: Proceedings of the 32nd ACM SIGKDD Conference on Knowledge Discovery and Data Mining}{August 9 -- 13, 2026}{Jeju, Korea}
\acmBooktitle{Proceedings of the 32nd ACM SIGKDD Conference on Knowledge Discovery and Data Mining (KDD'26), August 9--13, 2026, Jeju, Korea}
\acmPrice{15.00}
\acmISBN{978-1-4503-XXXX-X/18/06}

\begin{document}

%%
%% The "title" command has an optional parameter,
%% allowing the author to define a "short title" to be used in page headers.
\title{{\model}: Expanding Exploration for LLM Agents via Retrieval-Augmented Policy Optimization}

%%
%% The "author" command and its associated commands are used to define
%% the authors and their affiliations.
%% Of note is the shared affiliation of the first two authors, and the
%% "authornote" and "authornotemark" commands
%% used to denote shared contribution to the research.
\author{Siwei Zhang, Yun Xiong*, Xi Chen, Zi'an Jia, Renhong Huang, Jiarong Xu, Jiawei Zhang}
\email{swzhang24@m.fudan.edu.cn}
% \orcid{1234-5678-9012}
% \author{G.K.M. Tobin}
% \authornotemark[1]
% \email{webmaster@marysville-ohio.com}
\affiliation{
  \institution{Fudan University, Zhejiang University, UC Davis}
  \city{}
  \country{}
}

% \author{Lars Th{\o}rv{\"a}ld}
% \affiliation{
%   \institution{Fudan University}
%   \city{Hekla}
%   \country{Iceland}}
% \email{larst@affiliation.org}

% \author{Valerie B\'eranger}
% \affiliation{%
%   \institution{Inria Paris-Rocquencourt}
%   \city{Rocquencourt}
%   \country{France}
% }

% \author{Aparna Patel}
% \affiliation{%
%  \institution{Rajiv Gandhi University}
%  \streetaddress{Rono-Hills}
%  \city{Doimukh}
%  \state{Arunachal Pradesh}
%  \country{India}}

% \author{Huifen Chan}
% \affiliation{%
%   \institution{Tsinghua University}
%   \streetaddress{30 Shuangqing Rd}
%   \city{Haidian Qu}
%   \state{Beijing Shi}
%   \country{China}}

% \author{Charles Palmer}
% \affiliation{%
%   \institution{Palmer Research Laboratories}
%   \streetaddress{8600 Datapoint Drive}
%   \city{San Antonio}
%   \state{Texas}
%   \country{USA}
%   \postcode{78229}}
% \email{cpalmer@prl.com}

%%
%% By default, the full list of authors will be used in the page
%% headers. Often, this list is too long, and will overlap
%% other information printed in the page headers. This command allows
%% the author to define a more concise list
%% of authors' names for this purpose.
\renewcommand{\shortauthors}{Trovato and Tobin, et al.}

%%
%% The abstract is a short summary of the work to be presented in the
%% article.
\begin{abstract}
Agentic Reinforcement Learning (Agentic RL) has shown remarkable potential in large language model-based (LLM) agents.
These works can empower LLM agents to tackle complex tasks via multi-step, tool-integrated reasoning.
However, an inherent limitation of existing Agentic RL methods is their reliance on a pure \textit{on-policy} paradigm for exploration, restricting exploration to the agent's self-generated outputs and preventing the discovery of new reasoning perspectives for further improvement.
While recent efforts incorporate auxiliary off-policy signals to enhance exploration, they typically utilize full off-policy trajectories for \textit{trajectory-level} policy estimation, overlooking the necessity for the fine-grained, \textit{step-level} exploratory dynamics within agentic rollout. 
In this paper, we revisit exploration in Agentic RL and propose \textbf{\underline{R}}etrieval-\textbf{\underline{A}}ugmented \textbf{\underline{P}}olicy \textbf{\underline{O}}ptimization (\textbf{\model}), a novel RL framework that introduces retrieval to explicitly expand exploration during~training. To achieve this, we decompose the Agentic RL training process into two phases: (i) Hybrid-policy Agentic Rollout, and (ii) Retrieval-aware Policy Optimization.
Specifically, we propose a Hybrid-policy Agentic Rollout strategy, which allows the agent to continuously reason over the retrieved off-policy step-level traces. It dynamically extends the agent's reasoning receptive field, enabling broader exploration conditioned on external behaviors.
Subsequently, we introduce the Retrieval-aware Policy Optimization mechanism, which calibrates the policy gradient estimation with retrieval reward and~importance shaping, stabilizing training and prioritizing retrieval-illuminating exploration. Extensive experiments show that {\model} achieves an \textbf{+5.0\%} average gain on fourteen datasets across three agentic reasoning tasks, while delivering \textbf{1.2}x faster training~efficiency.
\end{abstract}


%% A "teaser" image appears between the author and affiliation
%% information and the body of the document, and typically spans the
%% page.
%%\begin{teaserfigure}
%%  \includegraphics[width=\textwidth]{sampleteaser}
%%  \caption{Seattle Mariners at Spring Training, 2010.}
%%  \Description{Enjoying the baseball game from the third-base
%%  seats. Ichiro Suzuki preparing to bat.}
%%  \label{fig:teaser}
%%\end{teaserfigure}

%%
%% This command processes the author and affiliation and title
%% information and builds the first part of the formatted document.
\maketitle

\section{Introduction}
\label{sec:intro}
Large Language Model-based (LLM) agents \cite{agent_1, agent_2, agent_3} have demonstrated expressive tool-integrated reasoning capabilities across a broad range of real-world tasks. Unlike traditional single-step reasoning \cite{skywork, kimi}, LLM agents invoke multi-step reasoning loops through iterative tool calls, enabling dynamic, interactive engagement with environments \cite{ToolStar}. Particularly noteworthy is how such capabilities have been realized through Agentic Reinforcement Learning (Agentic RL) \cite{agentic_survey}. 
These methods \cite{GiGPO, AEPO, Tree-GRPO} typically~adopt group-based algorithms, \eg, GRPO \cite{GRPO}, to optimize the step-level reasoning of agents, thus improving their task-solving~performance.

\begin{figure*}
    \centering
    \includegraphics[width=.99\linewidth]{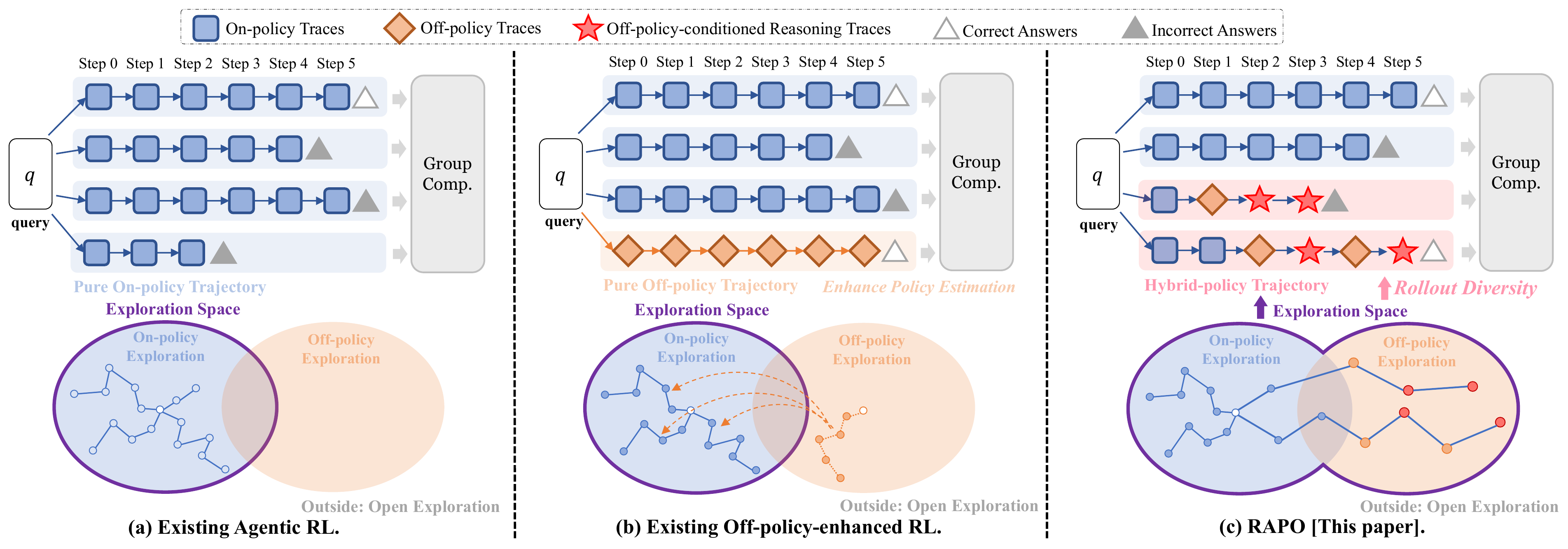}
    \vspace{-3mm}
    \caption{Comparison between existing methods and our framework. (a) Existing Agentic RL methods are inherently on-policy, resulting in a limited exploration space bounded by the native agent.
    (b) Off-policy-enhanced RL methods statically integrate full off-policy trajectories for trajectory-level policy estimation, failing to capture the dynamic, step-level exploration within agentic rollout.
    (c) Our {\model} introduces retrieval and allows the on-policy agent to continuously reason over the retrieved off-policy step-level traces, explicitly expanding its reasoning receptive field for exploration and thus increasing rollout~diversity.}
    \label{fig:intro}
    \vspace{-2mm}
\end{figure*}

Nonetheless, \textit{effective policy exploration} \cite{exploration_1, AEPO} remains an open challenge in Agentic RL, as it requires the policy to discover sufficiently diverse reasoning trajectories during rollout.
Toward this goal, recent Agentic RL methods propose rollout reconstruction, such as branching \cite{ARPO} or tree-search \cite{Tree-GRPO}, to enrich the candidate behaviors available for step-level exploration.
Although effective to some extent, a fundamental limitation inherent in these methods is their rigid reliance on an \textit{on-policy} exploration, where the model explores and learns exclusively from its self-generated reasoning trajectories across repeated trials. 
As shown in Fig.~\ref{fig:intro}(a), such a pure on-policy paradigm unconsciously constrains the global exploration space to the intrinsic behaviors of the base agent, leading to insufficient exploration for further performance improvement. 
This hypothesis is also supported by recent empirical findings \cite{on-policy_1, on-policy_2, on-policy_3}, which show that on-policy RL predominantly amplifies pre-existing behaviors within the base LLM, rather than uncovering new reasoning strategies beyond its native exploratory horizon.

To overcome the aforementioned exploration boundary, recent works \cite{LUFFY, ReplayBuffer} integrate auxiliary off-policy signals into RL training and exploit external reasoning behaviors to advance policy optimization.
However, as illustrated in Fig.~\ref{fig:intro}(b), these approaches leverage the entire off-policy trajectories solely for \textit{trajectory-level} group computation, overlooking the fine-grained, \textit{step-level} dynamics within agentic reasoning.
As a result, they merely strengthen the policy estimation over the already-observed rollouts in a static manner, failing to dynamically, explicitly enlarge exploration throughout the rollout process.
This observation inspires us to ask: \textit{Is it possible to explicitly inject off-policy signals into the step-level rollout process in Agentic RL, extending the agent's reasoning receptive field and thereby unlocking broader exploration?}

In this paper, we improve Agentic RL at the algorithmic level~and propose \textbf{\underline{R}}etrieval-\textbf{\underline{A}}ugmented \textbf{\underline{P}}olicy \textbf{\underline{O}}ptimization (\textbf{\model}), a novel RL framework that introduces retrieval to explicitly broaden exploration during training.
Beyond existing Agentic RL~methods that only generate pure on-policy trajectories, as depicted in Fig.~\ref{fig:intro}(c), {\model} incorporates a \textit{Hybrid-policy Agentic Rollout} strategy that produces hybrid-policy trajectories via off-policy-conditioned reasoning.
This design supports the agent to continuously reason over the retrieved off-policy step-level traces, substantially extending its reasoning receptive field and exposing the model to adaptively absorb these external behaviors for subsequent richer~exploration.

While incorporating external traces can facilitate exploration, it may inevitably introduce instability during policy optimization.
To tackle this issue, we further introduce a \textit{Retrieval-aware Policy Optimization} mechanism to calibrate the policy gradient estimation within {\model}.
Specifically, we design a Retrieval Reward that automatically encapsulates and quantifies the impact of retrieval within each rollout. This reward enables the agent to understand how retrieved information contributes to its reasoning, providing a principled signal for retrieval-aware exploration.
Meanwhile, we propose a Retrieval Importance Shaping that selectively prioritizes retrieval-informative tokens during optimization. It rebalances policy estimation using the retrieved-token proportion, encouraging the model to allocate greater attention to those externally-expanded behaviors.
Notably, {\model} applies retrieved token masking for stable optimization. As such, our framework can reduce the on-policy generation during agentic rollout while minimizing the number of gradient-bearing tokens within policy updates, ultimately leading to improved training efficiency.

In summary, our key contributions are as follows:
\vspace{-3mm}
\begin{itemize}[itemsep=0pt, parsep=0pt, leftmargin=*]
\item We address the challenge of exploration in Agentic RL and present {\model}. To the best of our knowledge, {\model} is the first Agentic RL framework that explicitly harnesses retrieval to augment the agent’s step-level exploration capabilities during training. 
\item We develop a Hybrid-policy Agentic Rollout to allow the agent to continuously reason over the retrieved off-policy traces, promoting expanded exploration and thus facilitating rollout diversity.
\item We propose a Retrieval-aware Policy Optimization, which calibrates policy gradient estimation through a carefully-designed retrieval reward and importance shaping, effectively improving training stability and overall effectiveness of {\model}.
\item Extensive experiments on fourteen benchmarks across three agentic reasoning tasks demonstrate that {\model} consistently outperforms baselines with 5.0\% average gains, while exhibiting markedly 1.2x faster training efficiency.
\end{itemize}

\section{Related Work}
\subsection{Agentic Reinforcement Learning}
Agentic Reinforcement Learning (Agentic RL) \cite{Agent_4, agent_5, agent_6, agent_9, agent_10} plays an important role for LLM agents, enabling them to interact with tool environments (\eg, python or search engines) \cite{SimpleTIR} and perform multi-step reasoning guided by feedback.
To effectively facilitate such tool-integrated reasoning, existing Agentic RL methods typically employ GRPO-based algorithms \cite{VerlTool, ToolRL, agent_8} to train agents to invoke tools for improved downstream problem-solving.
Despite these advances, effective policy exploration remains a fundamental challenge in Agentic RL \cite{agent_7, websailor, exploration_1}.
To this end, recent works attempt to address this limitation via rollout reconstruction, such as adaptive branching \cite{ARPO} or tree-search \cite{Tree-GRPO}, which significantly improve the exploration capabilities of agents during training.

While effective to some degree, these Agentic RL methods are inherently on-policy, restricting exploration to the native behaviors of the base agent. Instead, our {\model} allows the agent to reason over the retrieved off-policy step-level traces during rollout, explicitly expanding its reasoning receptive field for better exploration.

% explicitly harnesses step-level off-policy traces to dynamically expand the reasoning receptive field of agent, 
% diversify agent reasoning, promoting exploration beyond its native perspective.

\begin{figure*}
    \centering
    \includegraphics[width=.99\linewidth]{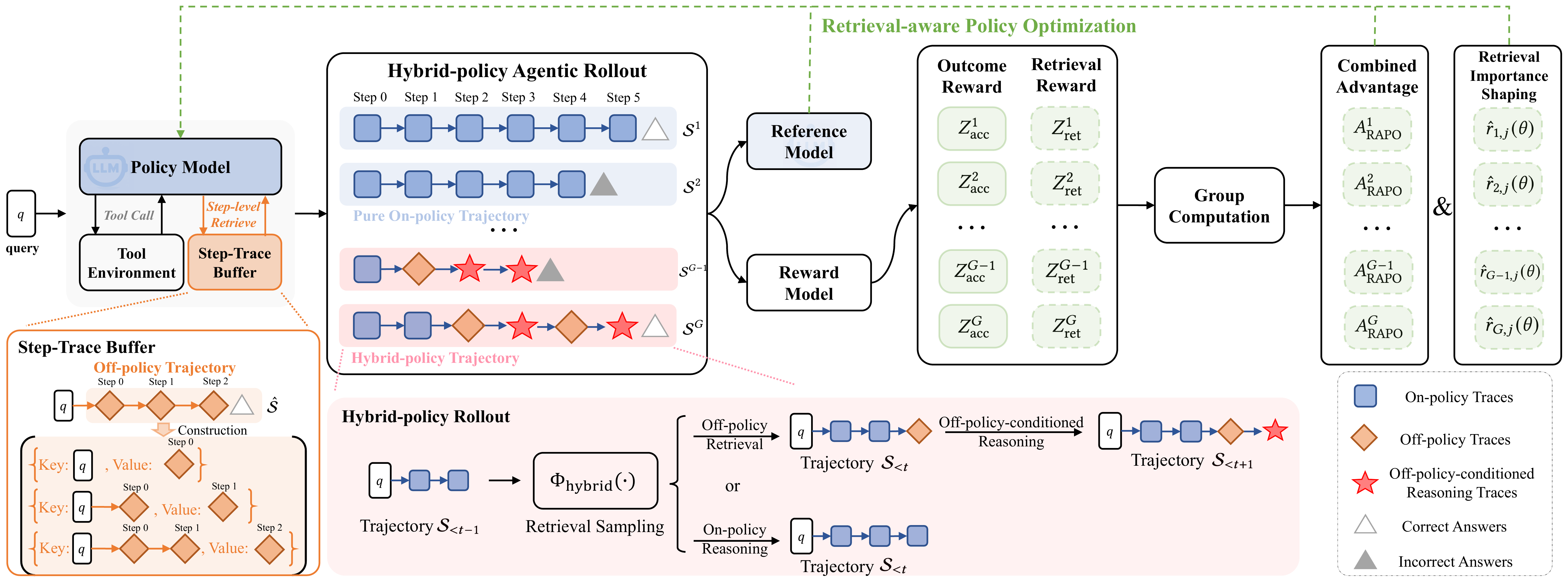}
    % \vspace{-3mm}
    \caption{Overview of the {\model}. 
    {\model} introduces a Hybrid-policy Agentic Rollout strategy that supports off-policy-conditioned reasoning, which enables the agent to receive the retrieved off-policy traces to broaden exploration beyond its intrinsic reasoning behaviors.
    Meanwhile, it incorporates a Retrieval-aware Policy Optimization mechanism with retrieval reward and importance shaping, ensuring effective and stable policy gradient estimation during training.
    }
    % \vspace{-2mm}
    \label{fig:method}
\end{figure*}

% \vspace{-5mm}
\subsection{On- and Off-Policy RL for LLM Training}
Based on how experience is utilized during policy optimization, RL for LLM training can be divided into two categories: on-policy and off-policy methods \cite{on-off-policy, on-off-policy_1, on-off-policy_2}. On-policy methods strictly update the policy through reasoning trajectories generated by the current policy LLM \cite{agent_12, agent_10, Resum}, ensuring training stability but potentially constraining the exploration space \cite{on-policy_1, on-policy_2, off_policy_1}. To mitigate this issue, recent works augment on-policy RL with off-policy signals to improve performance. They typically introduce an auxiliary LLM~\cite{LUFFY} or a replay buffer \cite{ReplayBuffer} to integrate off-policy trajectories for optimization, enabling effective exploration during~training.

The above methods focus on single-step reasoning and incorporate off-policy signals solely for trajectory-level policy~estimation. 

\noindent
In contrast, we integrate off-policy traces into the multi-step agentic rollout process, preserving step-level exploration dynamics while providing new insights during RL training.

\subsection{Entropy-Related RL for LLM Training}
Entropy \cite{Entropy_1, Entropy_2} serves as a well-established measure of model uncertainty, which has been widely used in LLM post-training \cite{agentic_survey, agentic_survey_2}. Recent works exploit entropy to monitor the reasoning states of LLMs during rollout \cite{Entropy_3} and incorporate it into policy optimization to improve performance \cite{Entropy_4}. In the context of Agentic RL, entropy has also been leveraged to enable adaptive branching during rollout~\cite{ARPO} and construct balanced supervisory signals during optimization~\cite{AEPO}, thus enhancing the tool-use capabilities of~agents.

Unlike these entropy-related works, we leverage entropy to quantify the contribution of retrieval to the agent's reasoning, allowing us to assess the retrieval-aware exploration for policy estimation.

\section{Preliminaries}
\label{sec:preliminaries}
\subsection{Problem Definition}\label{sec:problem_definition}
\textit{Definition 3.1.} \textbf{Multi-step Agentic Reasoning.}
Multi-step agentic reasoning performs iterative \textit{Thought-Action-Observation} circles between an LLM agent and an external tool environment for task solving. 
Given a query $q$, at each step $t = \{0, 1,\dots, T-1\}$, the agent $\pi_{\theta}$ generates a thought $\tau_t$ and an action $\alpha_t$ based on the current reasoning history. 
The action $\alpha_t$ is a parsable instruction that can be executed by tools such as \textit{python} or \textit{search} engines,~which then returns an observation $o_t$ as feedback.
We define the resulting reasoning trace at step $t$ as $s_t = \left(\tau_t, \alpha_t, o_t \right)$, and denote a complete $T$-step reasoning trajectory as $\mathcal{S}=\left(s_0, s_1, \cdots, s_{T-1}\right)$. We also use $\mathcal{S}_{<t} = \left(s_0, \dots, s_{t-1}\right)$ to represent the reasoning history before step~$t$.

\noindent
\textit{Definition 3.2.} \textbf{Agentic RL.}
Given a query $q$ sampled from the dataset $\mathcal{Q}$, an agent $\pi_{\theta}$, Agentic RL aims to maximize the expected reward of $T$-step agentic reasoning trajectory $\mathcal{S}\sim\pi_{\theta}(\cdot \mid q; \; T)$~by:
\begin{equation}
    \max_{\pi_{\theta}}\; \mathbb{E}_{q \sim \mathcal{Q}, \; \mathcal{S} \sim \pi_{\theta}(\cdot \mid q; \; T)} \Big[R(q, \mathcal{S}) \Big],
\end{equation}
where $R(\cdot)$ denotes a rule-based or model-based reward function~\cite{GRPO}.

\subsection{Entropy Computation for Step-level Traces}
Entropy provides a quantitative measure of uncertainty in LLM-generated token distributions, which has been widely adopted to monitor the real-time reasoning states of LLMs during training and inference \cite{Entropy_1, Entropy_2, Entropy_3, Entropy_4}. For the $i$-th token within the $t$-th step-level reasoning trace $s_t$, let $\textbf{p}_i \in \mathbb{R}^V$ denote its probability vector over the vocabulary size $V$, the token-level entropy is computed as:
\begin{equation}\label{eq:entropy}
    H_{s_t, i} = - \sum_{j=1}^{V} p_{i,j} \log p_{i,j}, 
    \; \text{where} \; \mathbf{p}_i = \mathrm{Softmax} \left(\mathbf{z}_i\right).
\end{equation}
Here, $\mathbf{z}_i \in \mathbb{R}^V$ represents the pre-softmax logits of the $i$-th token within $s_t$. The entropy of the step-level trace $s_t$ is then obtained by averaging all token-level entropies as $H_{s_t} = \mathrm{Mean}\left(\{H_{s_t, i}\}_{i=1}^{|s_t|}\right) \in \mathbb{R}$.

\section{Methodology}\label{sec:method}
As discussed earlier, the exploration capacity of existing Agentic RL methods is inherently bounded by the native agent. 
To tackle this issue, we propose {\model}, a novel Agentic RL framework that introduces retrieval to explicitly enlarge exploration during training. 
To achieve this, our {\model} consists of two key components: (i) a Hybrid-policy Agentic Rollout strategy that conducts dynamic off-policy-conditioned reasoning for step-level exploration expansion; and (ii) a Retrieval-aware Policy Optimization mechanism to ensure training stability and effectiveness. We will introduce these components in the following subsections.

% a Step-Trace Buffer that maintains step-level off-policy reasoning traces, (ii) a Hybrid-policy Agentic Rollout mechanism for promoting off-policy-conditioned reasoning , and (iii) a Retrieval-aware Policy Optimization procedure to ensure training stability.

% \vspace{-8mm}
\subsection{Hybrid-policy Agentic Rollout}
\label{sec:hybrid_policy_agentic_rollout}
In this section, we introduce Hybrid-policy Agentic Rollout, a strategy that retrieves the off-policy traces from a Step-Trace Buffer and seamlessly inserts them into the ongoing multi-step agentic rollout. 

% a Step-Trace Buffer, which stores the step-level off-policy reasoning traces to enable off-policy-conditioned reasoning during {\model} rollout. It will be described below.

\subsubsection{Step-Trace Buffer}\label{sec:step_trace_buffer}
Unlike existing works \cite{ReplayBuffer} that rely on \textit{trajectory-grained} buffers for experience replay, we instead construct a Step-Trace Buffer to record \textit{step-grained} reasoning traces collected from an off-policy agent. 
Concretely, for each query $q$ within the training dataset, we first employ an off-policy agent $\pi_{\theta_{\text{off}}}$ to generate $N$ independent multi-step reasoning trajectories ${\{\hat{\mathcal{S}}^{n}\}}_{n=1}^{N}$, each accompanied by its corresponding outcome reward.
To ensure buffer quality, we apply reward-aware filtering and retain only the top-$K$ reward-ranked trajectories. 
Each preserved trajectory $\hat{\mathcal{S}} = \left(\hat{s}_0, \dots, \hat{s}_{T-1}\right)$ is then decomposed into individual step-level traces, which are stored separately in the Buffer.

As illustrated in Fig.~\ref{fig:method}, the stored items can be organized as a series of \textit{step-level key-value pairs}, \ie, $\{\text{Key:} \; \hat{\mathcal{S}}_{<t}, \text{Value:} \; \hat{s}_t\}$, where $\hat{\mathcal{S}}_{<t}$ denotes the off-policy reasoning history before each step $t$, and $\hat{s}_t$ is the output off-policy trace at that step. 
Such a step-grained design enables the Buffer to capture localized, context-conditioned reasoning behaviors, rather than treating full trajectories as units.
Consequently, our Buffer maintains a large corpus of diverse and high-quality off-policy traces, which serve as reusable reasoning primitives for context-similarity retrieval in hybrid-policy~rollout.

\subsubsection{Retrieval from Step-Trace Buffer}
To explicitly broaden exploration during rollout, we enable the model to retrieve relevant off-policy traces from the Step-Trace Buffer and harness these external behaviors to dynamically extend the agent’s reasoning receptive field.
Specifically, when retrieval is triggered at the $t$-th step, we construct a retrieval query using the current on-policy reasoning history $\mathcal{S}_{<t}$. A standard RAG-based retrieval procedure \cite{Search-r1} is then performed over the Buffer, which returns the Value (\ie, $ \hat{s}_t$) whose Key (\ie, $\hat{\mathcal{S}}_{<t}$) is most aligned with the retrieval query (\ie, $\mathcal{S}_{<t}$):
\begin{equation}
    \hat{s}_t \sim \mathrm{Retrieve\left(\mathcal{S}_{<t}\right)}.
\end{equation}
Here, $\hat{s}_t$ is the retrieved off-policy step-level trace, where its associated off-policy input history is most similar to the current on-policy contexts.
We then concatenate $\hat{s}_t$ into the agent's working contexts, allowing the agent to continuously reason over this external trace in subsequent reasoning steps.
Notably, both retrieval and concatenation are performed at the step level, which enables off-policy signals to modulate the exploration dynamics within agentic rollout.

\subsubsection{Rollout Procedure}
Unlike existing works \cite{LUFFY} that integrate off-policy signals for \textit{static, trajectory-level} policy estimation, {\model} utilizes the retrieved off-policy traces to facilitate \textit{dynamic, step-level} rollout exploration. 
Concretely, for each query $q$, {\model} produces two sets of rollouts: $N_{\text{on}}$ pure on-policy trajectories, and $N_{\text{hybrid}}$ hybrid-policy trajectories, in which the off-policy traces are interleaved with on-policy reasoning. 
These hybrid-policy trajectories are produced via a \textit{Retrieval-then-Reasoning} process as follows:

\begin{enumerate}[label=\ding{\numexpr191+\arabic*}, leftmargin=*]
    \item \textbf{Initialization}: 
    For each query $q$, the on-policy agent first generates $N_{\text{hybrid}}$ first-step reasoning traces $s_0 \sim \pi_{\theta}(\cdot \mid q; \; t=0)$ as the initialization of reasoning contexts.
    \item \textbf{Retrieval Sampling}: 
    At each subsequent step ($t>0$), we introduce retrieval sampling to probabilistically decide whether the next reasoning trace is generated by the on-policy agent, \ie, $s_t \sim \pi_{\theta}(\cdot \mid \mathcal{S}_{<t}; \; t)$, or retrieved from the Step-Trace Buffer, \ie, $s_t \sim \mathrm{Retrieve}\left(\mathcal{S}_{<t}\right)$, through the following distribution:
    \begin{equation}\label{eq:retrieval_sampling}
    \Phi_{\mathrm{hybrid}}(s_t = a \mid \mathcal{S}_{<t}) =
    \begin{cases}
    0.5, & \text{if } t > 0, a \sim \pi_{\theta}(\cdot \mid \mathcal{S}_{<t}; \; t); \\
    0.5, & \text{if } t > 0, a \sim \mathrm{Retrieve}\left(\mathcal{S}_{<t}\right). \\
    \end{cases}
    \end{equation}
    \item \textbf{Off-policy-conditioned Reasoning}: 
    Once retrieval is triggered, the agent proceeds to reason conditioned on this retrieved trace, allowing external behaviors to impact its decisions.
\end{enumerate}
Such a hybrid-policy rollout enables the on-policy agent to dynamically absorb off-policy traces beyond its native reasoning perspectives. As a result, the agent's reasoning receptive field is substantially broadened by these external behaviors, promoting richer exploration during rollout.
Notably, while retrieval sampling introduces stochasticity for diversity, it may also yield trajectories that never trigger retrieval. Therefore, the number of hybrid trajectories per query is variable with the upper-bound of~$N_{\text{hybrid}}$.

% Such a retrieval sampling reflects that each step decides whether to generate a on-policy trace, $\theta_{\text{on}}\left(\mathcal{S}_{<t}\right)$, or retrieve a off-policy trace, $\mathrm{Retrieve}\left(\mathcal{S}_{<t}\right)$, based on distribution $\Phi_{\mathrm{hybrid}}(\cdot \mid \mathcal{S}_{<t})$.

\subsection{Retrieval-aware Policy Optimization}\label{sec:retrieval-aware_policy_optimization}
Incorporating external traces inevitably introduces noise into policy optimization, potentially leading to training instability and suboptimal performance.
To address this issue, we propose Retrieval-aware Policy Optimization, which leverages a retrieval reward and an importance shaping mechanism to regularize policy updates.

\subsubsection{Retrieval Reward}\label{sec:retrieval_reward}
Retrievals within {\model} are not guaranteed to always benefit the reasoning process of agent, as they inherently lack any explicit supervision or heuristics.
This makes it essential to assess the contribution of each retrieval to the agent's reasoning, so that the model can reliably discriminate between constructive guidance and misleading interference.
However, evaluating retrieval contribution is non-trivial and presents two key challenges: (i) \emph{how to quantify retrieval quality}, \ie, whether retrieval truly provides useful external behaviors; and (ii) \emph{how to ensure timely retrieval}, \ie, whether retrieval is triggered at stages when external information is actually needed.
To tackle these challenges, we draw inspiration from recent entropy-related RL methods \cite{Entropy_1, Entropy_3, Entropy_2}, which utilize entropy as a proxy for reasoning uncertainty in LLMs, enabling effective monitoring of rollout dynamics~\cite{ARPO} and providing invaluable supervision signals for optimization~\cite{AEPO}.
Motivated by this, we propose an entropy-based retrieval reward that jointly captures retrieval quality and timing as follows:
\begin{itemize}[itemsep=0pt, parsep=0pt, leftmargin=*]
\item \textbf{Retrieval Quality.}
Intuitively, if a retrieval successfully provides helpful external behaviors, it will reduce the model's uncertainty and make subsequent reasoning more confident. For a retrieved off-policy trace $\hat{s}_{t}$ at step $t$, we measure its impact on entropy reduction by comparing the entropy before and after retrieval:
\begin{equation}\label{eq:entropy_drop}
    H_{\Delta \hat{s}_{t}} = -\frac{H_{\hat{s}_{t+1}} - H_{\hat{s}_{t-1}}}{H_{\hat{s}_{t-1}}},
\end{equation}
where $H_{\hat{s}_{t-1}}$ and $H_{\hat{s}_{t+1}}$ denote the step-level entropy before and after retrieval, respectively. To enhance discrimination and smooth the score, we then apply a scaling factor with an activation as:
\begin{equation}
    g_{\hat{s}_{t}} = \mathrm{tanh} \left(2 \cdot H_{\Delta \hat{s}_{t}}  \right) \in \left(-1, 1\right).
\end{equation}
Here, a positive $g_{\hat{s}_{t}}$ indicates that the corresponding retrieval has reduced uncertainty and provided benefits for the ongoing reasoning, whereas a negative value suggests~misleading.
\item \textbf{Retrieval Timing.}
In {\model}, retrieval-conditioned reasoning aims to encourage the agent to generate more diverse rollouts. Hence, retrieval should ideally occur when the agent is exhibiting strong exploratory behavior, where its reasoning is more susceptible to impact from external traces.
High pre-retrieval entropy, $H_{\hat{s}_{t-1}}$, naturally provides such exploratory states. Consequently, we utilize $H_{\hat{s}_{t-1}}$ as a timing signal within retrieval reward. 
\end{itemize}
% By encouraging retrieval when the agent is more uncertain, we enable more effective exploration, expanding the model’s reasoning space and facilitating greater diversity in its exploration.

\noindent
Overall, for a retrieval triggered at step $t$, we define its reward as:
\begin{equation}\label{eq:retrieval_reward}
Z_{\mathrm{ret}}(\hat{s}_{t}) = g_{\hat{s}_{t}} \cdot H_{\hat{s}_{t-1}},
\end{equation}
where $g_{\hat{s}_{t}} \in (-1, 1)$ evaluates the quality of the retrieved trace and $H_{\hat{s}_{t-1}} > 0$ reflects the retrieval timing. In this way, retrievals that (i) reduce uncertainty and (ii) occur during high-uncertainty states tend to receive higher rewards, while misleading or poorly timed retrievals are more likely to obtain lower or negative rewards. We then average the retrieval rewards across all retrievals within each hybrid-policy trajectory and denote it as $Z_{\mathrm{ret}} = \mathrm{Mean}\left(Z_{\mathrm{ret}}(\hat{s}_{t})\right)$. We also assign $Z_{\mathrm{ret}} = 0$ for each pure on-policy trajectory.

\subsubsection{Retrieval Importance Shaping}\label{sec:retrieval_importance_shaping}
Integrating retrievals into RL training introduces an additional challenge, as these retrieved off-policy tokens do not support gradient backpropagation throughout the training process.
This results in \textit{sparse} gradient signals within hybrid-policy trajectories, leading to insufficient optimization for effectively capturing retrieval-aware exploration.
To mitigate this issue, we introduce a Retrieval Importance Shaping mechanism, which rebalances gradients by upweighting on-policy tokens generated under off-policy-conditioned reasoning. The key idea is to compensate the GRPO token-level importance sampling ratio \cite{GRPO} with the retrieved-token proportion, ensuring that the model allocates greater optimization focus to the sparse-gradient contexts within hybrid-policy trajectories.

Formally, given a query $q$ and a policy agent $\pi_\theta$, for its generated $j$-th token at step $t$, $s_{t, j}$, we reshape the GRPO importance sampling ratio $r_{t, j}(\theta)$ \cite{GRPO} using retrieved-token proportion $\mathcal{F}_{\mathrm{ret}}$:
\begin{equation}\label{eq:retrieval_importance_shaping}
    \hat{r}_{t, j}(\theta) = (1 + m \cdot \mathcal{F}_{\mathrm{ret}}) \cdot r_{t, j}(\theta),
\end{equation}
where $m > 0$ is a predefined hyper-parameter and $\mathcal{F}_{\mathrm{ret}}\in (0, 1)$ denotes the proportion of the retrieved tokens to the full trajectory length; 
$r_{t, j}(\theta) = \pi_{\theta}(s_{t, j}|q, \mathcal{S}_{<t}) / \pi_{\theta_{\mathrm{old}}}(s_{t, j}| q, \mathcal{S}_{<t})$ is the importance sampling ratio to calibrate the gradient based on policy gradient theory \cite{policy_thoery}, as the solutions are generated by the old policy $\pi_{\theta_{\mathrm{old}}}$ before the update. 
Note that $\mathcal{F}_{\mathrm{ret}} = 0$ for on-policy trajectories, and thus this mechanism introduces no bias in their policy updates. In doing so, the retrieval-driven exploration is effectively consolidated.

\subsubsection{Training Objective}
Now, we introduce the training objective of our {\model}. Similar to GRPO \cite{GRPO}, given a query $q$ and a group of rollouts $\{\mathcal{S}^{i}\}^G_{i=1}$, we compute the advantage of retrieval rewards as:
\begin{equation}\label{eq:retrieval_advantage}
    A_{\mathrm{ret}}^i = \frac{Z_\mathrm{ret}^i - \mathrm{Mean}\left(\{Z_\mathrm{ret}^i\}^G_{i=1}\right)}{\mathrm{Std}\left(\{Z_\mathrm{ret}^i\}^G_{i=1}\right)}.
\end{equation}
Then, we follow AEPO \cite{AEPO} and combine the retrieval advantage with the outcome advantage $A_{\mathrm{acc}}^i$ for each rollout by:
\begin{equation}\label{eq:combined_advantage}
    A_\mathrm{RAPO}^i = \left(1 + a \cdot A_{\mathrm{ret}}^i\right)\cdot A_{\mathrm{acc}}^i,
\end{equation}
where $a > 0$ is a predefined hyper-parameter.
Finally, the training objective of {\model} is defined as follows:
\begin{equation}\label{eq:final_loss}
\begin{aligned}
J_{\mathrm{RAPO}}(\theta) 
= \mathbb{E}&_{q \sim \mathcal{Q}, \, \mathcal{S} \sim \pi_{\mathrm{RAPO}}(\cdot \mid q)} 
\Bigg[
    \frac{1}{G} \sum_{i=1}^{G} \frac{1}{\lvert \mathcal{S}^i \rvert} 
    \sum_{j=1}^{\lvert \mathcal{S}^i \rvert}
    \min\Big( 
        \hat{r}_{i,j}(\theta)\,\hat{A}^i_{\mathrm{RAPO}}, \\
        & \mathrm{clip}\big(\hat{r}_{i,j}(\theta),\,1-\epsilon,\,1+\epsilon\big)
        \hat{A}^i_{\mathrm{RAPO}}
    \Big)
    - \beta\, \mathbb{D}_{\mathrm{KL}}
\Bigg],
\end{aligned}
\end{equation}
where $\mathrm{clip}(\cdot)$ clamps the importance ratio into $[1-\epsilon, 1+\epsilon]$ to ensure that the current policy is within the trust region \cite{clip}, and $\mathbb{D}_{\mathrm{KL}}$ is the KL divergence between the current and reference policies~\cite{GRPO}.

\section{Experiments}

\subsection{Experimental Settings}\label{sec:settings}
\subsubsection{Datasets}\label{sec:datasets}
We conduct experiments with \textbf{14 datasets} across three multi-step agentic reasoning tasks: (i) \textit{Computational Reasoning}, including GSM8K \cite{GSM8K}, MATH \cite{MATH}, MATH500 \cite{MATH500}, AIME2024, and AIME2025\footnote{\url{https://huggingface.co/datasets/AI-MO/aimo-validation-aime}}; (ii) \textit{Knowledge-Intensive Reasoning}, including WebWalkerQA \cite{webWalker}, HotpotQA \cite{HotpotQA}, 2WikiMultihopQA \cite{2wiki}, Musique \cite{Musique}, and Bamboogle \cite{Bamboogle}; and (iii) \textit{Web-Agentic Reasoning}, including SimpleQA \cite{Bamboogle}, GAIA \cite{GAIA}, WebWalkerQA \cite{webWalker}, and BrowseComp \cite{Browsecomp}. Due to page limitations, details of these datasets are provided in Sec.~\ref{app_sec:datasets} of the Appendix. All dataset splits follow the standard settings adopted in existing Agentic RL methods~\cite{ARPO, Tree-GRPO, Search-r1}.

\vspace{-1.5mm}
\subsubsection{Baselines}\label{sec:baselines}
For comparison, we select \textbf{13 baselines} from three families: (i) \textit{Tool-Integrated Reasoning Methods}, including Search-o1 \cite{Search-o1}, Search-R1 \cite{Search-r1}, and ToolStar \cite{ToolStar}; (ii) \textit{Off-policy Learning Methods}, including SFT, RolloutReplay \cite{ReplayBuffer}, and LUFFY \cite{LUFFY}; and (iii) \textit{Agentic RL Methods}, including Single-step RL (GRPO \cite{GRPO}, DAPO \cite{DAPO}, and GPPO \cite{GPPO}) and Multi-step RL (GiGPO \cite{GiGPO}, Tree-GRPO \cite{Tree-GRPO}, ARPO \cite{ARPO}, and AEPO \cite{AEPO}). Descriptions of all baselines are put in Sec.~\ref{app_sec:baselines}. For Agentic RL Methods, we evaluate using three representative LLM backbones, including Qwen2.5-3B-instruct \cite{Qwen}, Llama3-8B-instruct \cite{Llama}, and Qwen2.5-7B-instruct~\cite{Qwen}. For our {\model}, we use AEPO-Qwen3-14B \cite{AEPO} as the default off-policy agent. The results on other off-policy models are also provided in Sec.~\ref{sec:robustness}.

\vspace{-1.5mm}
\subsubsection{Training and Evaluation}
For Computational Reasoning and Knowledge-Intensive Reasoning, we choose the widely-used RL training dataset from Tool-Star \cite{ToolStar}.
In these two settings, we employ \textbf{Python} and \textbf{Search} tools, with search results provided by a local search server \cite{Search-r1} built on a Wikipedia dump \cite{Wikipedia}. For~Web-Agentic Reasoning, we follow the configuration of Tree-GRPO \cite{Tree-GRPO} and employ real search APIs during both training and evaluation. 

\vspace{-1.5mm}
\subsubsection{Evaluation Metrics}
We follow ARPO \cite{ARPO} and adopt F1 scores as the evaluation metric in both Knowledge-Intensive and Web-Agentic Reasoning. For Computational Reasoning, we also follow ARPO \cite{ARPO} to employ LLM-as-Judge to evaluate answers. We use Pass@1 as the metric. Implementation details are put in Sec.~\ref{app_sec:settings}.

\begin{table*}
\caption{Results (\%) for Computational Reasoning and Knowledge-Intensive Reasoning tasks. Unless specified, the baselines use Qwen2.5-7B-Instruct as the backbone. The best results are highlighted in bold, and the second-best results are \underline{underlined}.}
\label{tab:main_results}
\setlength{\tabcolsep}{4.1pt}
\vspace{-3mm}
\begin{tabular}{ccccccccccccc}
\toprule
\multicolumn{2}{c}{\multirow{2}{*}{\cellcolor{tabularColor}\textbf{Methods}}} & \multicolumn{5}{c}{\cellcolor{tabularColor}\textbf{Computational Reasoning}} & \multicolumn{5}{c}{\cellcolor{tabularColor}\textbf{Knowledge-Intensive Reasoning}} & \multirow{2}{*}{\cellcolor{tabularColor}\textbf{Avg.}} \\
\cmidrule(lr){3-7} \cmidrule(lr){8-12}
\multicolumn{2}{c}{\cellcolor{tabularColor}} & \cellcolor{tabularColor} AIME24 & \cellcolor{tabularColor} AIME25 & \cellcolor{tabularColor} MATH500 & \cellcolor{tabularColor} GSM8K & \cellcolor{tabularColor} MATH & \cellcolor{tabularColor} WebWalker & \cellcolor{tabularColor} HQA & \cellcolor{tabularColor} 2Wiki. & \cellcolor{tabularColor} MuSiQ. & \cellcolor{tabularColor} Bamb. & \cellcolor{tabularColor} \\
\midrule
\multicolumn{13}{c}{\textbf{Tool-Integrated Reasoning Methods}} \\ 
\midrule
\multicolumn{2}{c}{Search-o1}   & 6.7  & 10.0 & 61.8 & 80.2 & 73.6 & 10.4 & 22.1 & 21.8 & 5.4  & 32.0 & \cellcolor{avg} 32.4 \\
\multicolumn{2}{c}{Search-R1}   & 16.7 & 6.7  & 63.8 & 82.4 & 81.2 & 12.0 & 26.9 & 25.9 & 16.2 & 40.4 & \cellcolor{avg} 37.2 \\
\multicolumn{2}{c}{Tool-Star}   & 30.0 & 26.7 & 77.2 & 89.4 & 85.6 & 18.5 & 38.1 & 40.8 & 14.9 & 41.5 & \cellcolor{avg} 46.3 \\ 
\midrule
\multicolumn{13}{c}{\textbf{Off-policy Learning Methods}} \\ 
\midrule
\multicolumn{2}{c}{SFT}          & 12.2 & 17.3 & 53.8 & 77.9 & 77.3 & 10.9 & 46.0 & 36.9 & 17.2 & 39.7 & \cellcolor{avg} 38.9 \\
\multicolumn{2}{c}{RolloutReplay}& 10.8 & 18.0 & 55.0 & 76.9 & 75.0 & 11.1 & 44.8 & 30.7 & 19.8 & 38.9 & \cellcolor{avg} 38.1 \\
\multicolumn{2}{c}{LUFFY}        & 29.4 & 23.1 & 75.2 & 83.2 & 80.5 & 14.7 & 46.7 & 37.0 & 18.0 & 40.6 & \cellcolor{avg} 44.8 \\ 
\midrule
\multicolumn{13}{c}{\textbf{RL Methods}} \\ 
\midrule
\multirow{10}{*}{\rotatebox{90}{\textbf{Qwen2.5-3B-Instruct}}} & \multicolumn{12}{l}{\textbf{\textit{Single-step RL}}} \\
& GRPO  & 20.0 & 13.3 & 72.0 & 86.0 & 81.0 & 7.3  & 39.0 & 36.3 & 15.2 & 36.8 & \cellcolor{avg} 40.7 ($\Delta_{\text{base}}$) \\
& DAPO  & 20.0 & 16.7 & 71.2 & 85.0 & 81.2 & 6.8  & 37.2 & 36.1 & 16.2 & 35.9 & \cellcolor{avg} 40.6 \textcolor{down}{($\downarrow 0.1$)} \\
& GPPO & 17.2 & 19.5 & 69.7 & 86.3 & 80.5 & 13.9 & 40.0 & 36.9 & 16.5 & 36.9 & \cellcolor{avg} 41.7 \textcolor{up}{($\uparrow 1.0$)} \\
\cdashline{2-13}
& \multicolumn{12}{l}{\textbf{\textit{Multi-step Agentic RL}}} \\
& GiGPO & 21.8 & 19.8 & 70.1 & \underline{86.4} & 80.7 & 12.4 & 25.0 & 40.0 & 18.2 & 40.5 & \cellcolor{avg} 41.5 \textcolor{up}{($\uparrow 0.8$)} \\
& Tree-GRPO & 19.8 & 19.6 & 70.7 & 85.7 & 80.6 & 13.8 & \underline{42.4} & \underline{43.7} & 17.8 & \underline{43.2} & \cellcolor{avg} \underline{43.7} \textcolor{up}{($\uparrow 3.0$)} \\
& ARPO  & \underline{23.3} & \underline{20.0} & \underline{71.4} & 85.0 & \underline{82.5} & 12.9 & 37.9 & 41.1 & 17.1 & 38.9 & \cellcolor{avg} 43.0 \textcolor{up}{($\uparrow 2.3$)} \\
& AEPO  & 21.8 & 20.0 & 70.8 & 84.6 & 80.8 & \underline{15.8} & 36.1 & 43.2 & \underline{18.7} & 40.0 & \cellcolor{avg} 43.2 \textcolor{up}{($\uparrow 2.5$)} \\
& \ours{Ours} & \ours{\textbf{24.5}} & \ours{\textbf{24.8}} & \ours{\textbf{72.0}} & \ours{\textbf{87.2}} & \ours{\textbf{82.8}} & \ours{\textbf{18.0}} & \ours{\textbf{45.8}} & \ours{\textbf{48.9}} & \ours{\textbf{20.5}} & \ours{\textbf{45.9}} & \cellcolor{avg} \ours{\textbf{47.0}} \textbf{\textcolor{up}{($\uparrow 6.3$)}} \\
\cmidrule(l){2-13}
\multirow{10}{*}{\rotatebox{90}{\textbf{Llama3-8B-Instruct}}} & \multicolumn{12}{l}{\textbf{\textit{Single-step RL}}} \\
& GRPO  & 13.3 & 13.3 & 62.4 & 87.4 & 79.2 & 9.9  & 40.5 & 35.9 & 20.9 & 42.9 & \cellcolor{avg} 40.6 ($\Delta_{\text{base}}$) \\
& DAPO  & 16.7 & 13.3 & 61.2 & 87.4 & 76.4 & 18.5 & 36.9 & 37.9 & \textbf{28.8} & 41.2 & \cellcolor{avg} 41.8 \textcolor{up}{($\uparrow 1.2$)} \\
& GPPO  & 16.7 & 6.7  & 61.8 & 86.6 & 79.4 & 18.0 & 44.4 & 37.6 & 19.2 & \underline{48.6} & \cellcolor{avg} 41.9 \textcolor{up}{($\uparrow 1.3$)} \\
\cdashline{2-13}
& \multicolumn{12}{l}{\textbf{\textit{Multi-step Agentic RL}}} \\
& GiGPO & 20.0 & 13.3 & 62.4 & 87.4 & 77.2 & 22.1 & 42.8 & 38.9 & 20.1 & 47.9 & \cellcolor{avg} 43.2 \textcolor{up}{($\uparrow 2.6$)} \\
& Tree-GRPO & 22.1 & 14.2 & 64.2 & 86.1 & 78.1 & \underline{25.7} & 47.9 & \underline{41.1} & 26.1 & 47.8 & \cellcolor{avg} 45.3 \textcolor{up}{($\uparrow 4.7$)} \\
& ARPO  & 23.3 & \underline{16.7} & 64.6 & \textbf{88.0} & 80.2 & 23.8 & \textbf{48.9} & 40.2 & 25.9 & 48.1 & \cellcolor{avg} \underline{46.0} \textcolor{up}{($\uparrow 5.4$)} \\
& AEPO  & \underline{26.7} & 16.7 & \underline{65.8} & 87.6 & \underline{80.6} & 25.2 & 43.2 & 39.2 & 20.1 & 46.3 & \cellcolor{avg} 45.1 \textcolor{up}{($\uparrow 4.5$)} \\
& \ours{Ours} & \ours{\textbf{27.1}} & \ours{\textbf{17.4}} & \ours{\textbf{66.6}} & \ours{\underline{87.7}} & \ours{\textbf{80.8}} & \ours{\textbf{28.0}} & \ours{\underline{48.7}} & \ours{\textbf{41.4}} & \ours{\underline{28.1}} & \ours{\textbf{49.8}} & \cellcolor{avg} \ours{\textbf{47.6}} \textbf{\textcolor{up}{($\uparrow 7.0$)}} \\
\cmidrule(l){2-13}
\multirow{10}{*}{\rotatebox{90}{\textbf{Qwen2.5-7B-Instruct}}} & \multicolumn{12}{l}{\textbf{\textit{Single-step RL}}} \\
& GRPO  & 23.3 & 26.7 & 78.0 & \textbf{92.8} & 87.8 & 13.0 & 42.5 & 40.7 & 19.1 & 43.2 & \cellcolor{avg} 46.7 ($\Delta_{\text{base}}$) \\
& DAPO  & 20.0 & 23.3 & \underline{80.4} & 91.0 & 88.8 & 16.1 & 38.9 & 25.8 & \textbf{27.0} & 40.1 & \cellcolor{avg} 45.1 \textcolor{down}{($\downarrow 1.6$)} \\
& GPPO  & 26.7 & 23.3 & 76.2 & 91.6 & 87.6 & 22.9 & 42.9 & 40.1 & 21.8 & 46.2 & \cellcolor{avg} 47.9 \textcolor{up}{($\uparrow 1.2$)} \\
\cdashline{2-13}
& \multicolumn{12}{l}{\textbf{\textit{Multi-step Agentic RL}}} \\
& GiGPO & 30.0 & 20.0 & 78.4 & 91.6 & 87.6 & 21.7 & 39.7 & 38.4 & 21.1 & 45.1 & \cellcolor{avg} 47.4 \textcolor{up}{($\uparrow 0.7$)} \\
& Tree-GRPO & 31.0 & \textbf{30.0} & 79.2 & 90.1 & 88.5 & 21.0 & 44.6 & 42.3 & 20.2 & 44.0 & \cellcolor{avg} 49.1 \textcolor{up}{($\uparrow 2.4$)} \\
& ARPO  & 30.0 & 30.0 & 78.8 & 92.2 & 88.8 & 20.1 & 42.1 & \underline{42.8} & 21.8 & \underline{46.8} & \cellcolor{avg} 49.3 \textcolor{up}{($\uparrow 2.6$)} \\
& AEPO  & \underline{33.3} & 30.0 & 80.4 & 92.2 & \underline{90.0} & \underline{23.3} & \underline{45.8} & 37.7 & 18.8 & 44.2 & \cellcolor{avg} \underline{49.6} \textcolor{up}{($\uparrow 2.9$)} \\
& \ours{Ours} & \ours{\textbf{34.2}} & \ours{\underline{29.4}} & \ours{\textbf{81.1}} & \ours{\underline{92.7}} & \ours{\textbf{91.5}} & \ours{\textbf{24.9}} & \ours{\textbf{46.8}} & \ours{\textbf{42.9}} & \ours{\underline{22.1}} & \ours{\textbf{47.6}} & \cellcolor{avg} \ours{\textbf{51.3}} \textbf{\textcolor{up}{($\uparrow 4.6$)}} \\
\bottomrule
\end{tabular}
\vspace{-2mm}
\end{table*}

\begin{table*}
\caption{Results (\%) for Web-Agentic Reasoning tasks. The best results are highlighted in bold, and the second-best are \underline{underlined}.}
\label{tab:web_agent}
\setlength{\tabcolsep}{4.5pt}
\vspace{-3mm}
\begin{tabular}{ccccccccccccc}
\toprule
\multicolumn{2}{c}{\multirow{2}{*}{\cellcolor{tabularColor}\textbf{Methods}}} & \multirow{2}{*}{\cellcolor{tabularColor} \textbf{SimpleQA}} & \multicolumn{4}{c}{\cellcolor{tabularColor} \textbf{General AI Assistant}} & \multicolumn{4}{c}{\cellcolor{tabularColor} \textbf{WebWalkerQA}} & \multirow{2}{*}{\cellcolor{tabularColor} \textbf{Browse.}} & \multirow{2}{*}{ \cellcolor{tabularColor}\textbf{Avg.}} \\
\cmidrule(lr){4-7} \cmidrule(lr){8-11}
\multicolumn{2}{c}{\cellcolor{tabularColor}} & \cellcolor{tabularColor}  & \cellcolor{tabularColor} \textbf{Lv.1} & \cellcolor{tabularColor} \cellcolor{tabularColor} \textbf{Lv.2} & \cellcolor{tabularColor} \cellcolor{tabularColor} \textbf{Lv.3} & \cellcolor{tabularColor} \textbf{Overall} & \cellcolor{tabularColor} \textbf{Easy} & \cellcolor{tabularColor} \textbf{Med.} & \cellcolor{tabularColor} \textbf{Hard} & \cellcolor{tabularColor} \textbf{Overall} & \cellcolor{tabularColor} & \cellcolor{tabularColor} \\
\midrule
\multirow{2}{*}{\rotatebox{90}{\textbf{Direct}}}
& Qwen2.5-32B     & 7.7 & 8.8 & 7.7 & 3.0 & 7.6 & 6.2 & 9.4 & 5.8 & 7.4 & 2.2 & \cellcolor{avg} 6.6 \\
& DeepSeek-R1-32B & 12.6 & 19.2 & 7.8 & 4.1 & 11.7 & \underline{9.4} & \underline{13.3} & 9.4 & 11.0 & 2.4 & \cellcolor{avg} 10.1 \\
\midrule
\multirow{6}{*}{\rotatebox{90}{\textbf{Agentic RL}}}
& GRPO      & 61.5 & 17.7 & 14.9 & 4.5 & 14.7 & 8.9 & 11.4 & 11.6 & 10.9 & 2.3 & \cellcolor{avg} 15.1 ($\Delta_{\text{base}}$)\\
& GiGPO     & 61.8 & 18.3 & 17.3 & 3.2 & 13.3 & 8.0 & 10.8 & 9.9 & 11.8 & 2.7 & \cellcolor{avg} 14.2 \textcolor{down}{($\downarrow 0.9$)}\\
& Tree-GRPO & 62.4 & 19.3 & 17.5 & \underline{5.7} & 16.8 & 9.3 & 11.8 & \underline{11.9} & 11.2 & 2.7 & \cellcolor{avg} 15.3 \textcolor{up}{($\uparrow 0.2$)}\\
& ARPO      & \underline{63.8} & 19.7 & 17.7 & 5.2 & 15.8 & 9.8 & 10.6 & 10.8 & \underline{12.1} & 2.6 & \cellcolor{avg} 15.3 \textcolor{up}{($\uparrow 0.2$)}\\
& AEPO      & 62.1 & \textbf{20.8} & \underline{18.9} & 5.4 & \underline{16.9} & 9.0 & 11.7 & 10.4 & 11.6 & \underline{2.9} & \cellcolor{avg} \underline{15.9} \textcolor{up}{($\uparrow 0.8$)}\\
& \ours{Ours} & \ours{\textbf{64.7}} & \ours{\underline{20.6}} & \ours{\textbf{19.6}} & \ours{\textbf{6.0}} & \ours{\textbf{17.8}} & \ours{\textbf{9.8}} & \ours{\textbf{14.7}} & \ours{\textbf{12.9}} & \ours{\textbf{13.8}} & \ours{\textbf{4.4}} & \cellcolor{avg} \ours{\textbf{17.0}} \ours{\textcolor{up}{($\uparrow 1.9$)}}\\
\bottomrule
\end{tabular}
\vspace{-2mm}
\end{table*}

\subsection{Main Results}\label{sec:main_results}
\textbf{Computational and Knowledge-Intensive Reasoning.}
The results for Computational and Knowledge-Intensive Reasoning tasks are presented in Tab.~\ref{tab:main_results}.
Clearly, {\model} consistently improves the performance of all three LLM backbones across all datasets and tasks. Meanwhile, it achieves new state-of-the-art performance over the strongest baseline. These results validate the effectiveness of {\model}, which introduces retrieval to facilitate effective policy exploration in Agentic RL.
Although existing off-policy learning methods are suboptimal, {\model} still performs well. This highlights the importance of our hybrid-policy rollout strategy, which actively promotes fine-grained, step-level exploration dynamics within agentic rollout.
Moreover, Agentic RL tends to exhibit better stability than single-step RL, emphasizing the necessity of multi-step considerations in tool-integrated reasoning.

% explicit off-policy-guided reasoning during rollout, which helps the LLM to expand its exploration space and thus produce responses beyond native perspectives.

\noindent
\textbf{Web-Agentic Reasoning.}
We further evaluate {\model} using real-world web search APIs in Tab.~\ref{tab:web_agent}. We employ Qwen2.5-7B-instruct as the LLM backbone. From these results, we can see that our proposed {\model} still achieves the best performance across all baselines, highlighting its practicality and generalization for Web-Agentic QA.
Additionally, the overall performance gains observed on Web-Agentic Reasoning tasks are relatively modest compared to other tasks. We infer that this is due to the inherent difficulty of the evaluation data and the unavoidable failures of API calls.

\begin{figure}
  \centering
  %--- A 独占一行 ---
  \begin{subfigure}{0.5\linewidth}
  \hspace{-5mm}
    \includegraphics[width=1.1\linewidth]{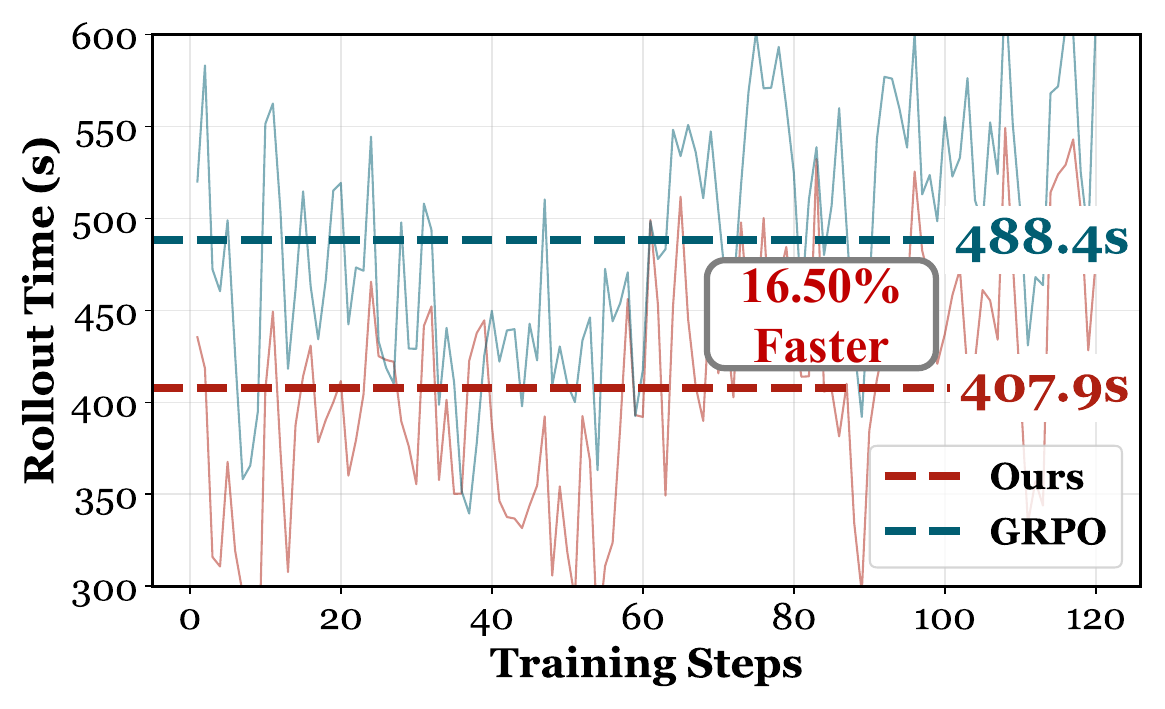}
    \vspace{-6mm}
    \caption{Rollout time.}
    \label{PlaceHolder}
  \end{subfigure}
  \hspace{-2mm}
  \begin{subfigure}{0.5\linewidth}
    \includegraphics[width=1.1\linewidth]{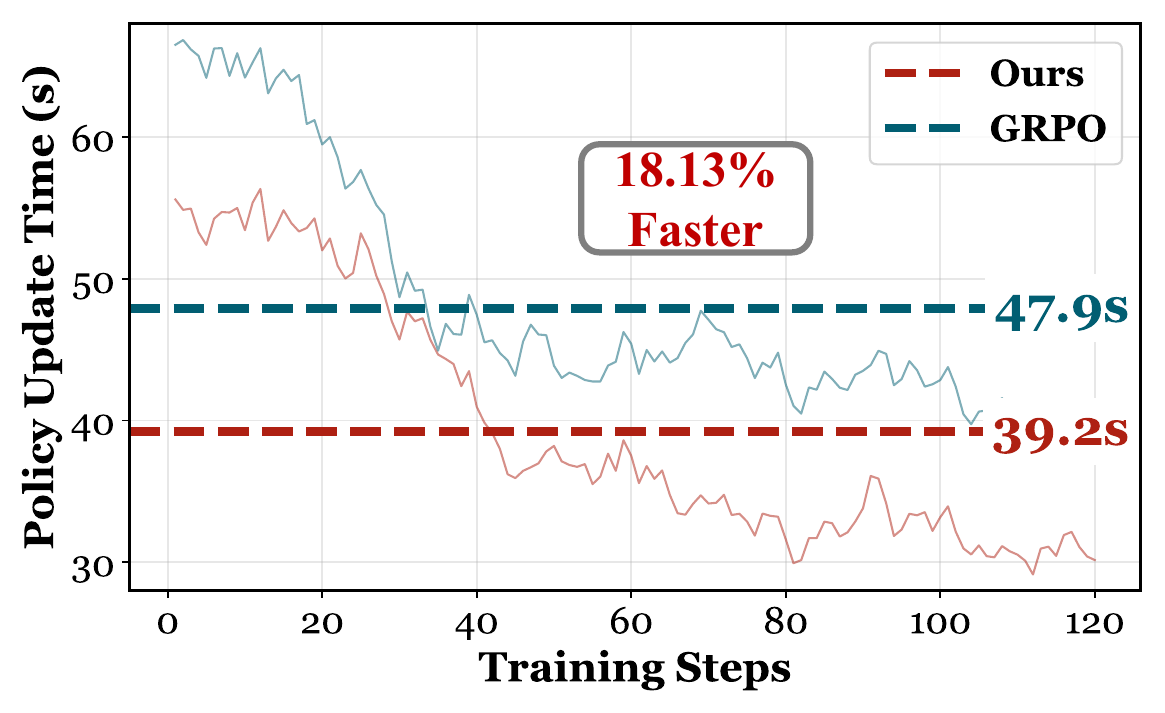}
    \vspace{-6mm}
    \caption{Policy update time.}
    \label{fig:PlaceHolder}
  \end{subfigure}
  \begin{subfigure}{0.5\linewidth}
  \hspace{-5mm}
    \includegraphics[width=1.1\linewidth]{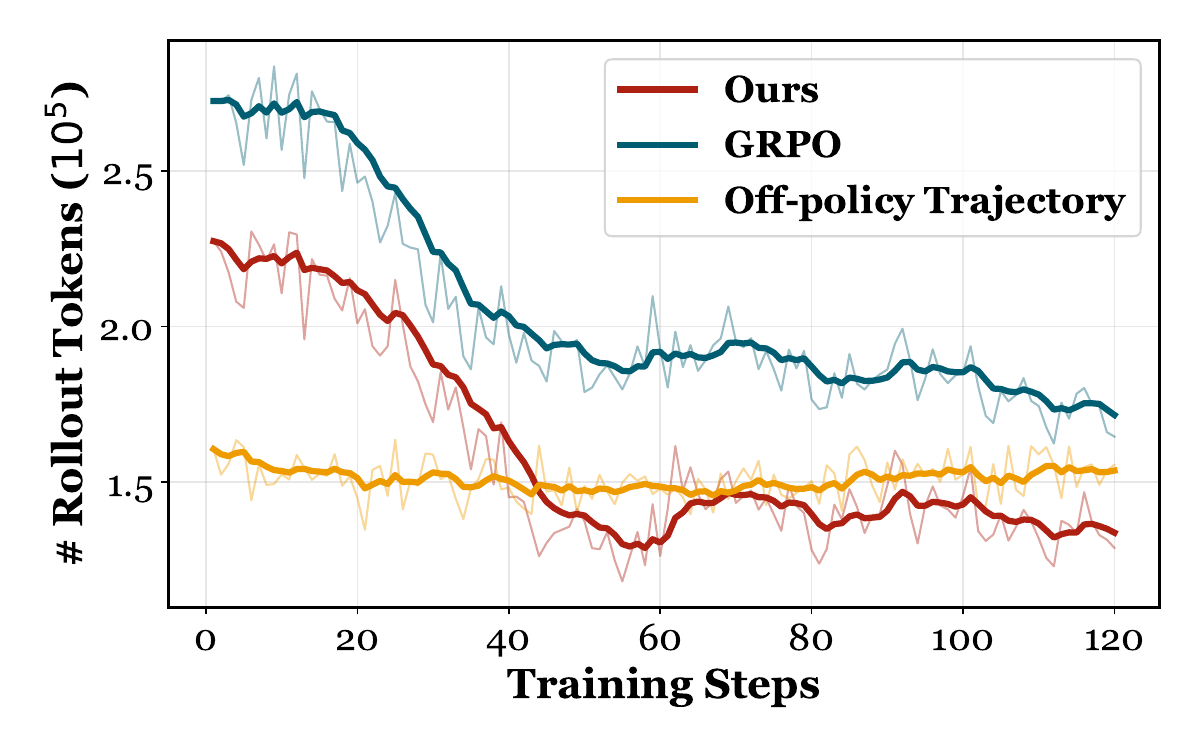}
    \vspace{-6mm}
    \caption{Rollout tokens.}
    \label{PlaceHolder}
  \end{subfigure}
  \hspace{-2mm}
  \begin{subfigure}{0.5\linewidth}
    \includegraphics[width=1.1\linewidth]{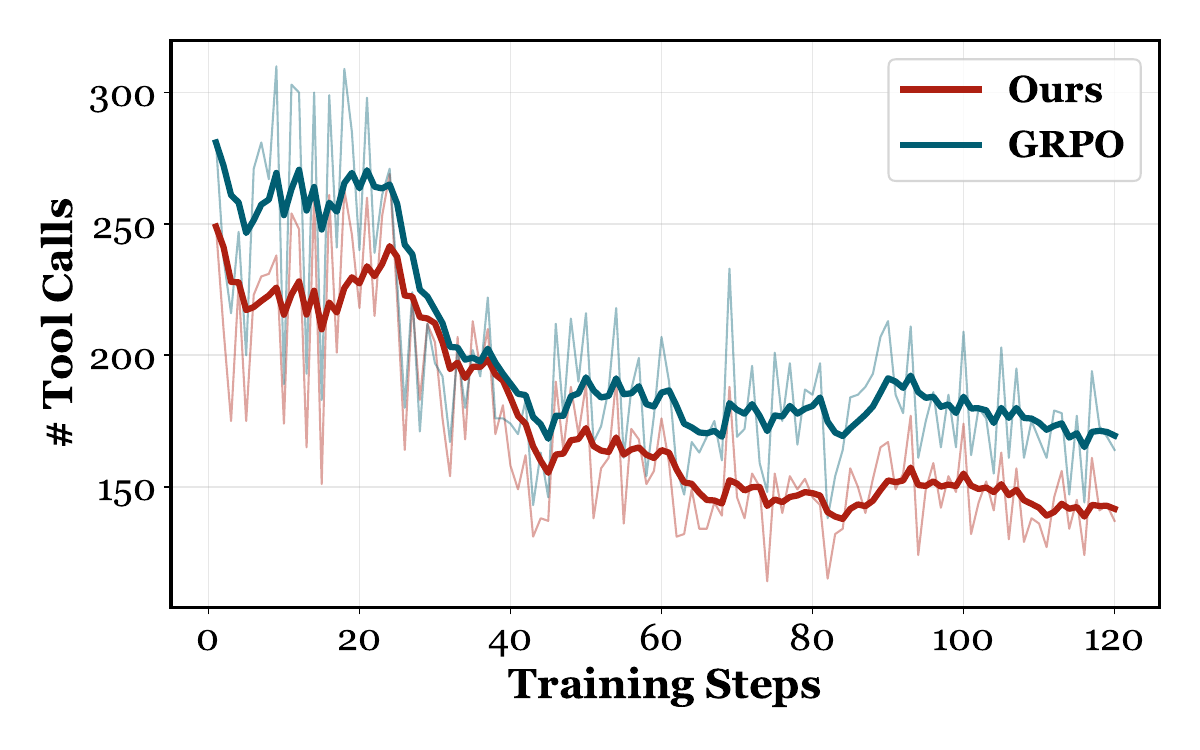}
    \vspace{-6mm}
    \caption{Tool calls.}
    \label{fig:PlaceHolder}
  \end{subfigure}
  \vspace{-8mm}
  \caption{Efficiency study. {\model} exhibits clear training efficiency in rollout time, policy update time, the number of rollout tokens, and the number of tool calls.}
  \vspace{-3mm}
  \label{fig:effiency}
\end{figure}

\subsection{Efficiency Study}\label{sec:efficiency}
In this section, we conduct an efficiency study to empirically validate the additional efficiency benefits from {\model} training. Specifically, we record several efficiency metrics at each RL training step, including rollout time, policy update time, the number of produced rollout tokens, and the number of tool calls. These experiments are performed on 4 NVIDIA RTX A100 (80 GB) GPUs.

The results are reported in Fig.~\ref{fig:effiency}. {\model} exhibits clear advantages in training efficiency over GRPO.
(i) Rollout time is substantially reduced, as retrieval alleviates the need for exhaustive on-policy generation at the rollout stage.
(ii) Policy updates are also accelerated via retrieved token masking, which explicitly reduces gradient-bearing tokens during optimization. 
(iii) {\model} produces fewer rollout tokens, whose distributions are more consistent with off-policy trajectories. This observation proves that the agent has successfully internalized and learned from off-policy reasoning patterns.
(iv) {\model} makes fewer tool calls, suggesting that the retrieved traces can supplement external knowledge for the agent and thus reduce its reliance on tool calls.

\begin{figure}
  \centering
  %--- A 独占一行 ---
  \begin{subfigure}{\linewidth}
    \centering
    \includegraphics[width=\linewidth]{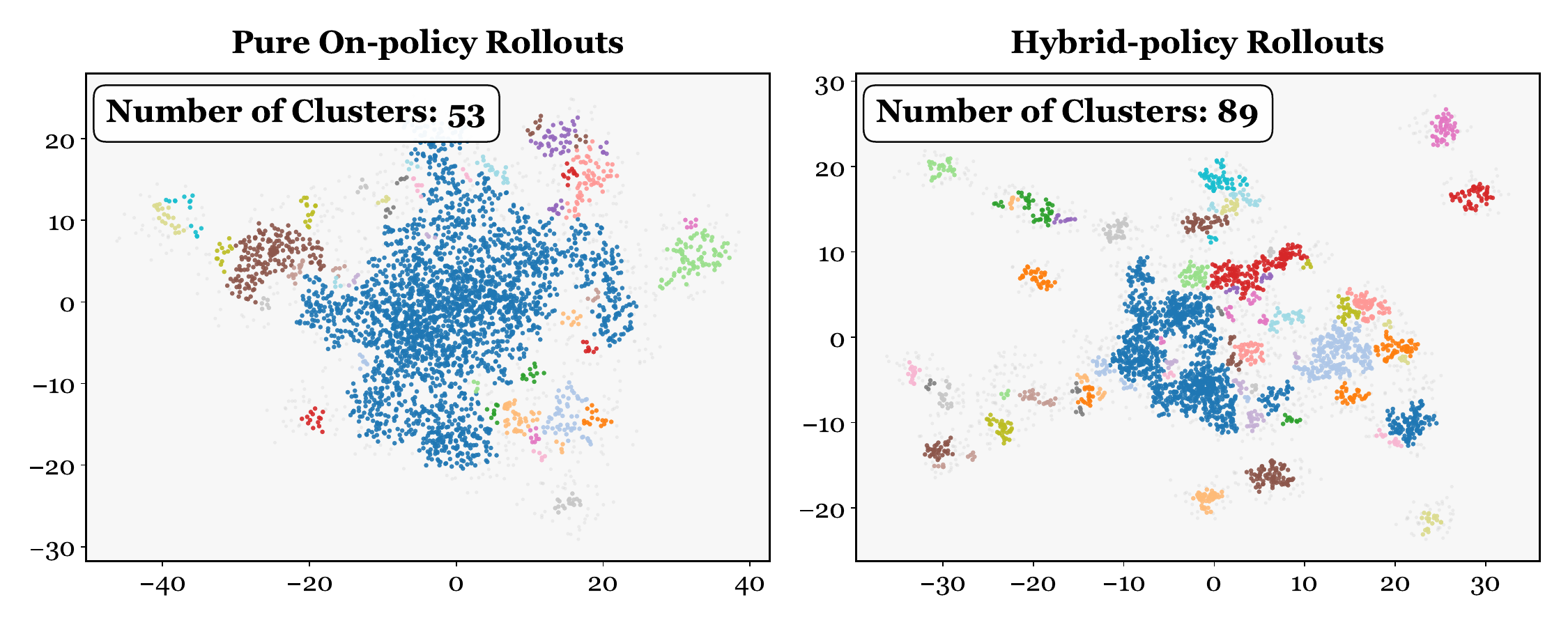}
    \vspace{-6mm}
    \caption{Rollout diversity.}
    \label{fig:comparison_hybrid_cluster}
  \end{subfigure}
  \begin{subfigure}{0.5\linewidth}
  \hspace{-5mm}
    \includegraphics[width=1.1\linewidth]{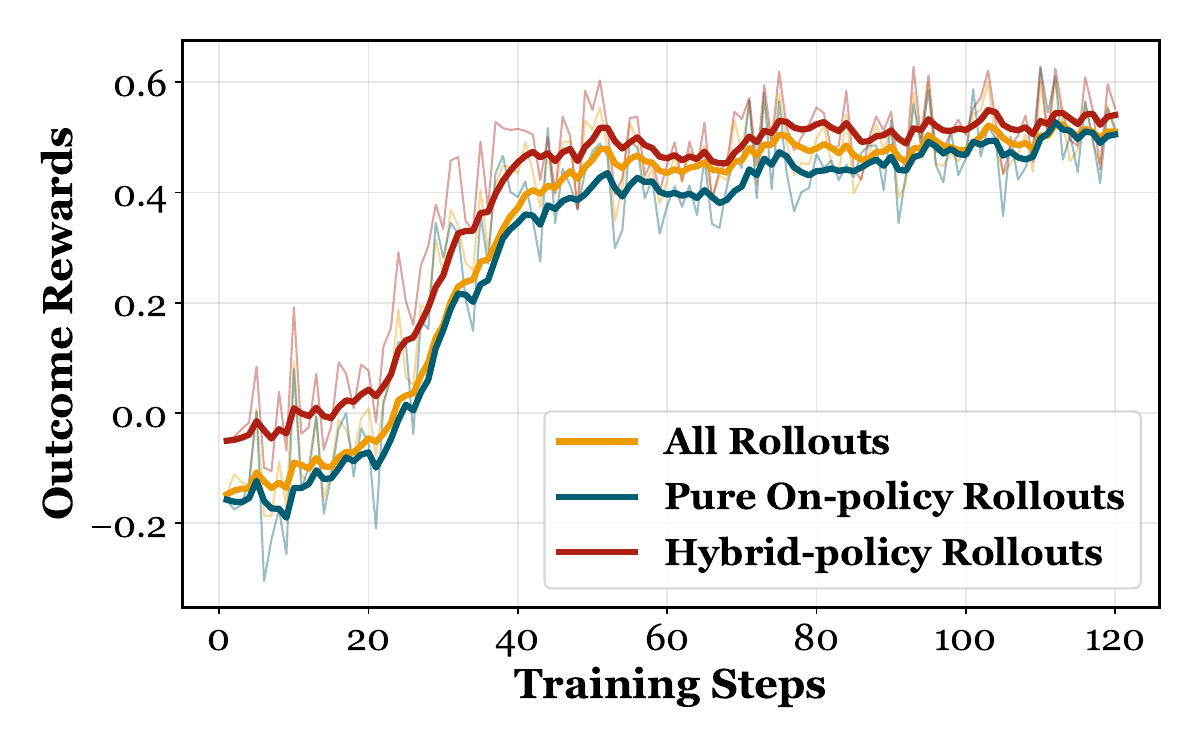}
    \vspace{-6mm}
    \caption{Outcome rewards.}
    \label{fig:comparison_hybrid_token}
  \end{subfigure}
  \hspace{-2mm}
  \begin{subfigure}{0.5\linewidth}
    \includegraphics[width=1.1\linewidth]{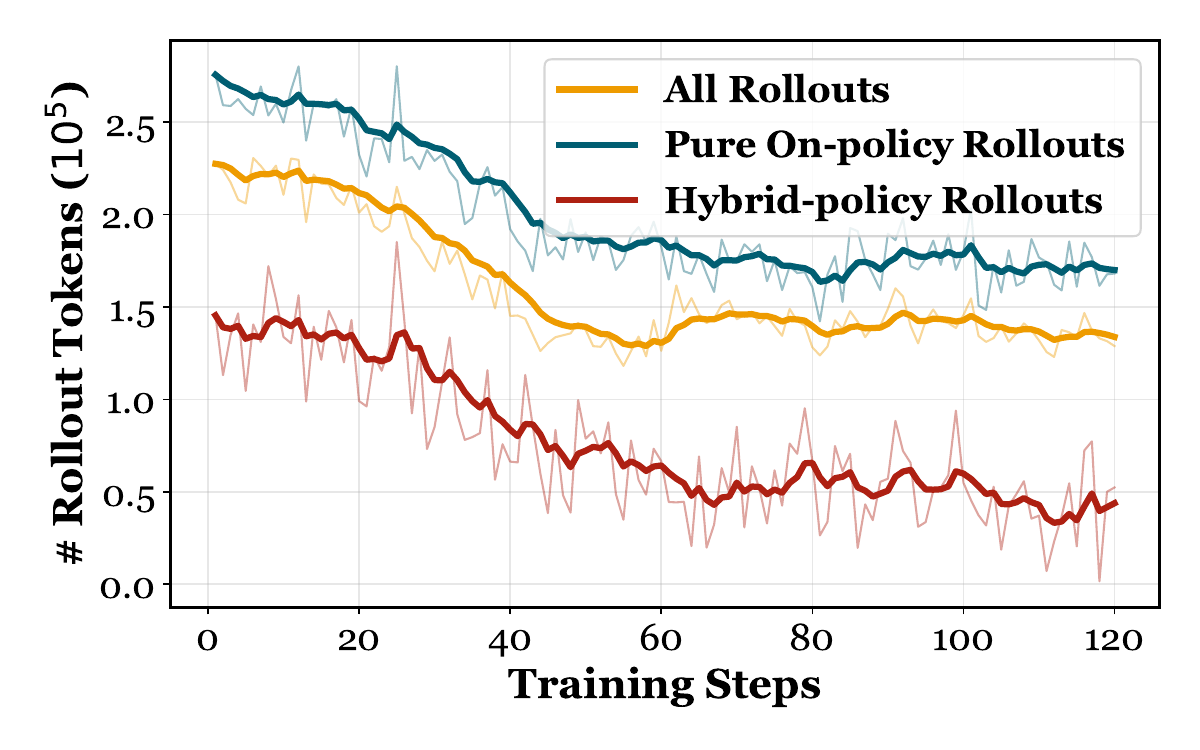}
    \vspace{-6mm}
    \caption{Rollout tokens.}
    \label{fig:comparison_hybrid_reward}
  \end{subfigure}
  \vspace{-8mm}
  \caption{Comparison analysis between pure on-policy rollouts and hybrid-policy rollouts.}
  \label{fig:comparison_hybrid}
\end{figure}

\subsection{Benefits of Retrieval}\label{sec:off-policy}
To understand how retrieval influences policy behaviors during rollout, we separately analyze \textit{pure on-policy rollouts} and \textit{hybrid-policy rollouts}, where we visualize their diversity, outcome rewards, and the number of rollout tokens, respectively.
To quantify diversity, we randomly sample approximately $7\mathrm{k}$ rollouts from each group and extract their semantic embeddings using BGEM3 \cite{BGEM3}. We then apply PCA and perform DB-SCAN \cite{DB-SCAN} clustering for visualization.

We present the results in Fig.~\ref{fig:comparison_hybrid}.
First, hybrid-policy rollouts form more clusters within semantic space, reflecting significantly higher rollout diversity. This can be attributed to the retrieval from external behaviors, which pushes exploration beyond the confined distribution of pure on-policy rollouts. 
Second, hybrid-policy rollouts achieve higher outcome rewards. This suggests that retrieval can improve reasoning quality, which provides clearer group advantages and facilitates better credit assignment during training. 
Third, hybrid-policy rollouts exhibit shorter trajectory lengths. This implies that retrieval can prevent redundant reasoning by~injecting informative signals, thus allowing the agent to arrive at solutions with fewer rollout tokens.

\begin{figure}
  \centering
  \begin{subfigure}{0.5\linewidth}
    \includegraphics[width=1.01\linewidth]{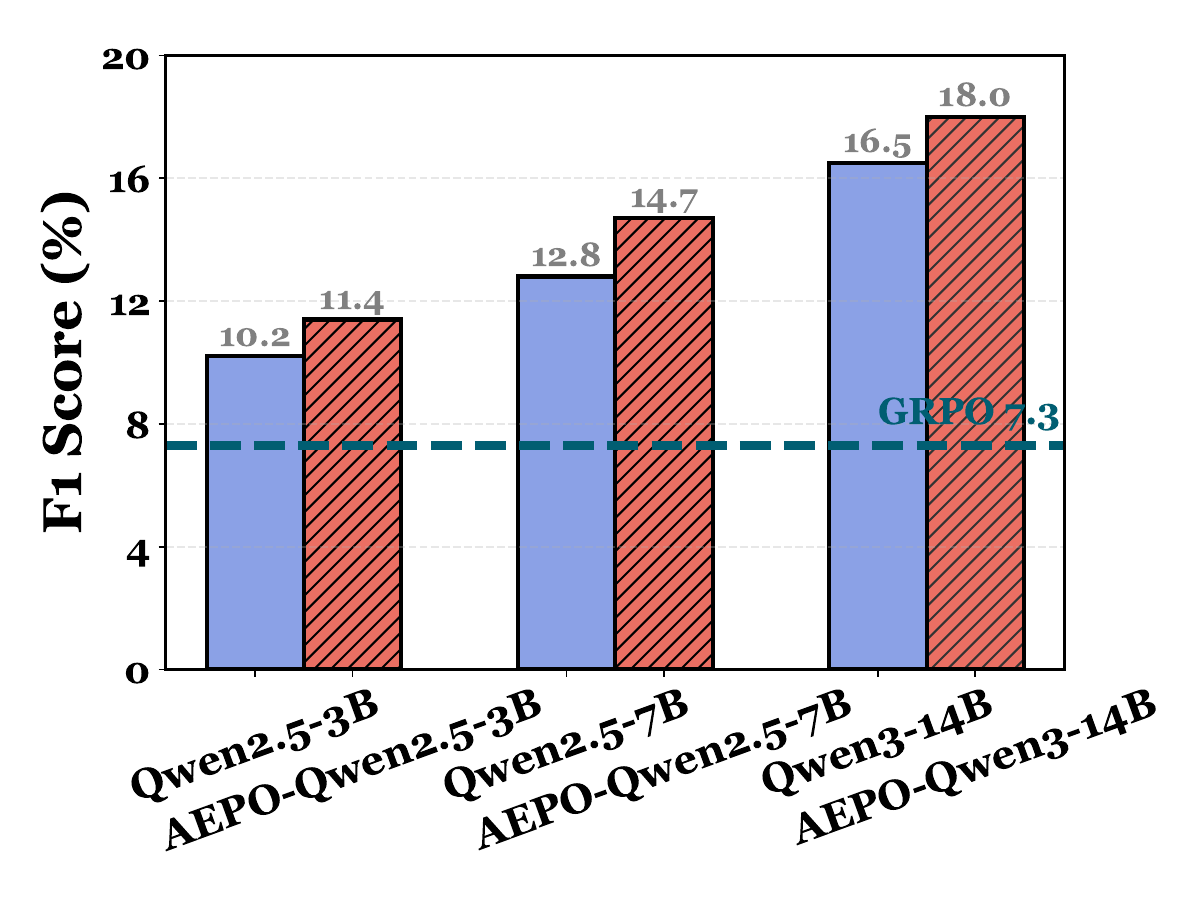}
    \vspace{-8mm}
    \caption{WebWalker.}
    \label{fig:robust_off_policy_model_webwalker}
  \end{subfigure}
  \hspace{-2mm}
  \begin{subfigure}{0.5\linewidth}
    \includegraphics[width=1.01\linewidth]{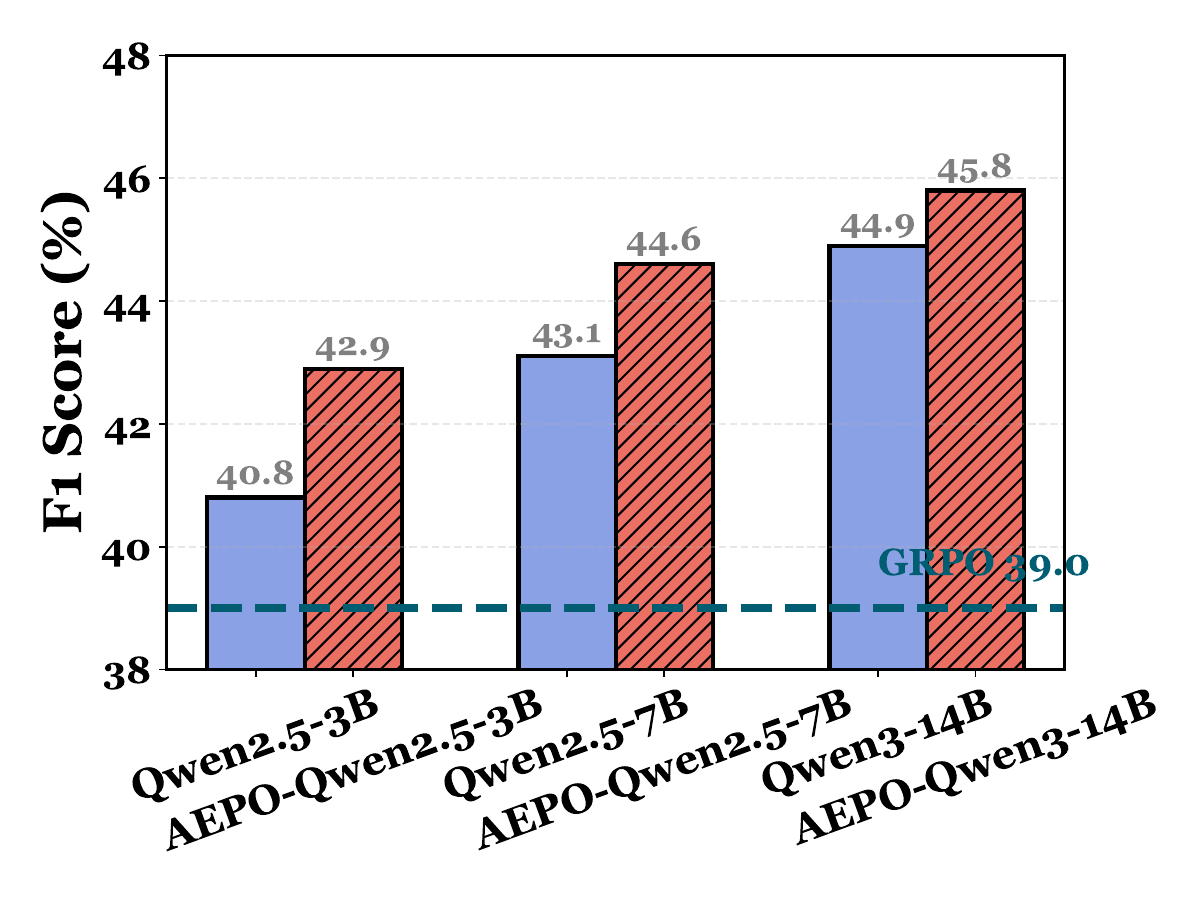}
    \vspace{-8mm}
    \caption{HQA.}
    \label{fig:robust_off_policy_model_HQA}
  \end{subfigure}
  \vspace{-8mm}
  \caption{Robustness study for different off-policy models.}
  \label{fig:robust_off_policy_model}
\end{figure}
\begin{figure}
  \centering
  \vspace{-4mm}
  \begin{subfigure}{0.5\linewidth}
    \includegraphics[width=1.01\linewidth]{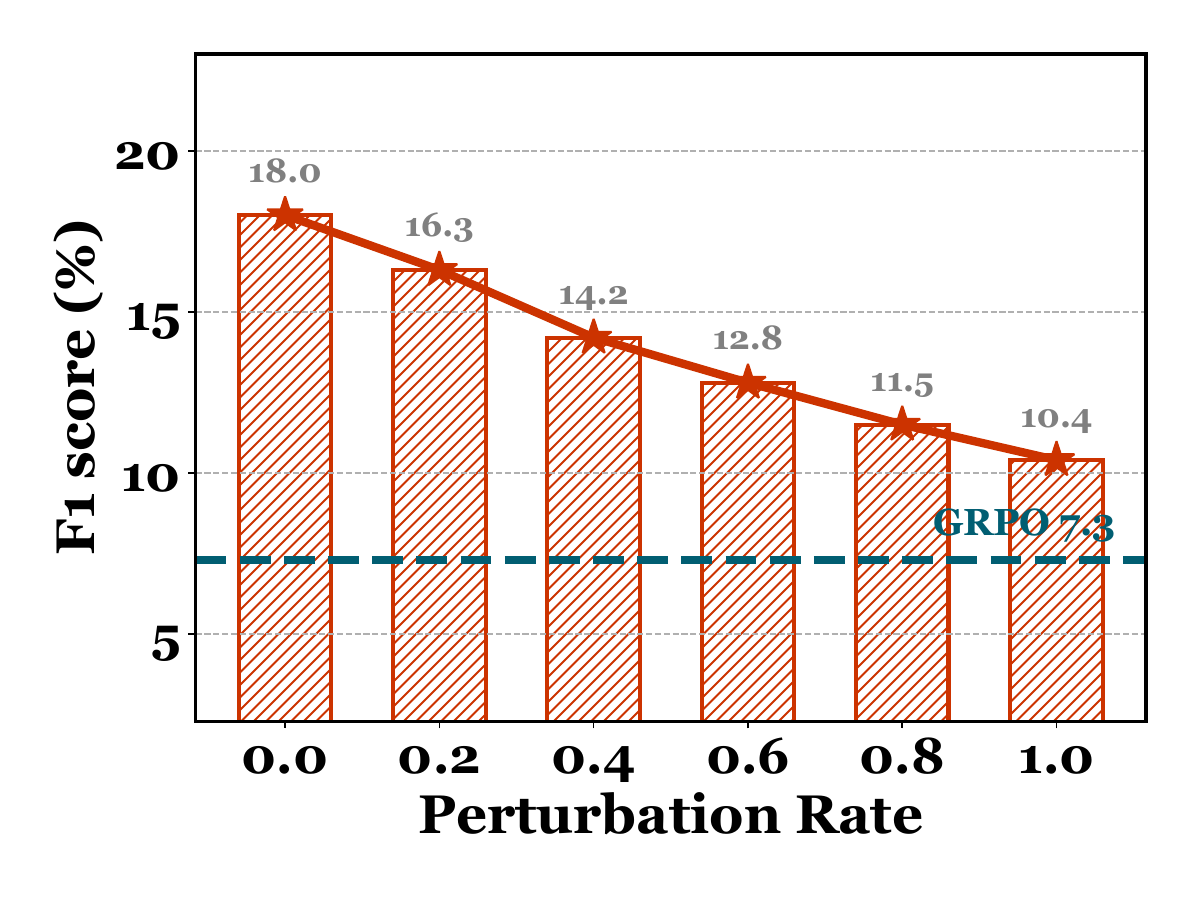}
    \vspace{-8mm}
    \caption{WebWalker.}
    \label{fig:robust_noise_webwalker}
  \end{subfigure}
  \hspace{-2mm}
  \begin{subfigure}{0.5\linewidth}
    \includegraphics[width=1.01\linewidth]{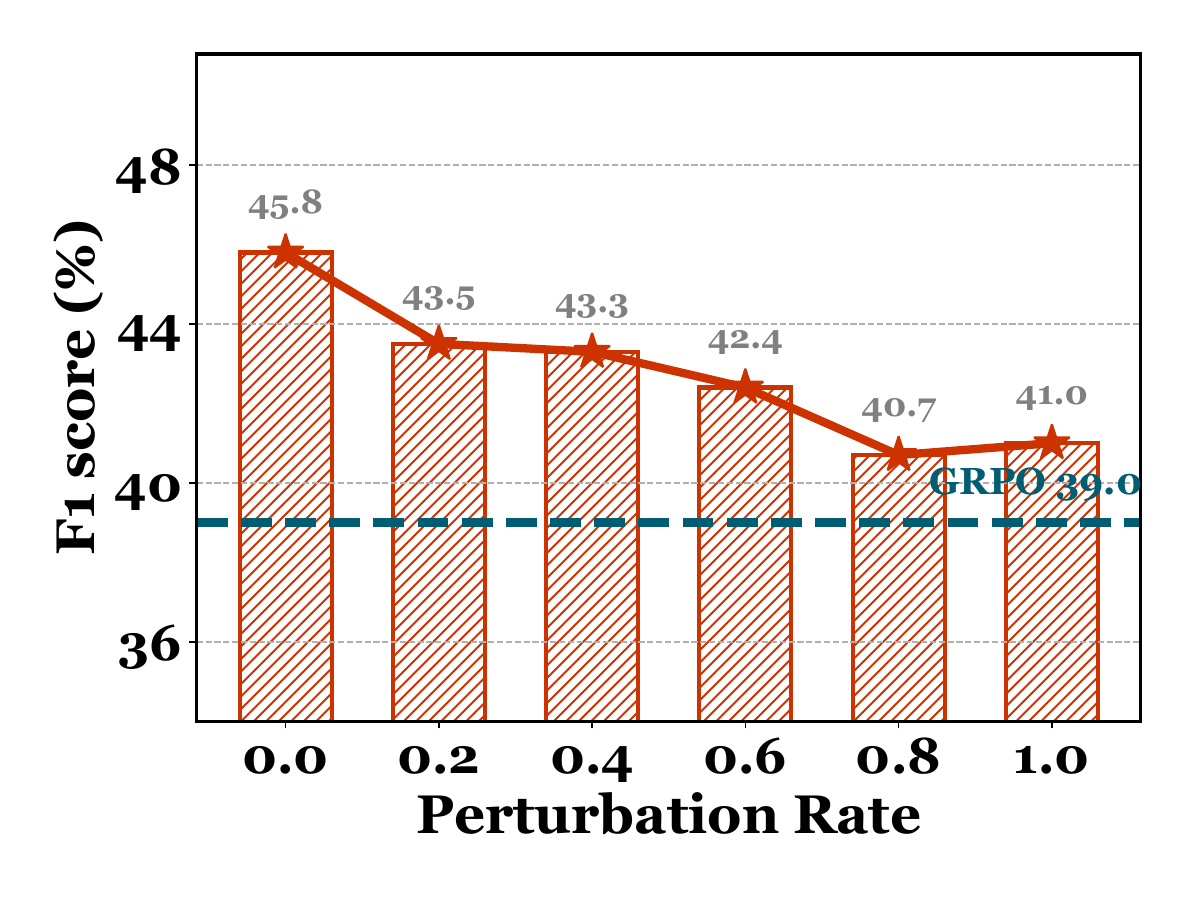}
    \vspace{-8mm}
    \caption{HQA.}
    \label{fig:robust_noise_HQA}
  \end{subfigure}
  \vspace{-8mm}
  \caption{Robustness study for noisy retrieval.}
  \label{fig:robust_noise}
\end{figure}

\subsection{Robustness Study for Retrieval}\label{sec:robustness}
\subsubsection{Robustness for Different Off-policy Models}
To evaluate the impact of buffer quality, we conduct a robustness study using various off-policy models for buffer construction, including AEPO-Qwen3-14B, AEPO-Qwen2.5-7B, AEPO-Qwen2.5-3B \cite{AEPO}, Qwen3-14B, Qwen2.5-7B, and Qwen2.5-3B \cite{Qwen}. 
As shown in Fig.~\ref{fig:robust_off_policy_model}, 
{\model} consistently outperforms GRPO, demonstrating its strong robustness under various off-policy models.
This can be due to our retrieval-aware optimization mechanism, which successfully estimates the retrieval from different buffer quality and thus maintains stability. 
It should also be noted that {\model} benefits from stronger off-policy models, producing better performance under a higher-quality~buffer.

% enables the policy estimation to adaptively extract and filter useful information from 

\vspace{-1.5mm}
\subsubsection{Robustness for Noisy Retrieval}
We further conduct the robustness study using noisy retrieval, where we introduce retrieval noise by replacing the retrieved trace with a randomly sampled trace at varying perturbation rates $p \in \{0.0, 0.2, 0.4, 0.6, 0.8, 1.0\}$. As depicted in Fig.~\ref{fig:robust_noise}, {\model} still exhibits strong robustness and consistently outperforms GRPO across all perturbation rates. Remarkably, {\model} surpasses GRPO even when the retrieval is entirely random ($p=1.0$). This indicates that the model can self-regulate by estimating the contribution of retrieved information. The results under $p=1.0$ also suggest that {\model} remains effective even when relevant queries are absent during buffer construction.

% automatically filtering out noisy retrievals when query-relevant off-policy traces are absent.

\begin{table}
\caption{Ablation study for retrieval reward.}
\label{tab:retrieval_reward}
\setlength{\tabcolsep}{1pt}
\vspace{-3mm}
\begin{tabular}{cccccc}
\toprule
\cellcolor{tabularColor} & \multirow{2}{*}{\cellcolor{tabularColor}\textbf{w/o RR}} & \multicolumn{2}{c}{\cellcolor{tabularColor}$H^\text{Low}$} & \multicolumn{2}{c}{\cellcolor{tabularColor}$H^\text{High}$} \\ \cmidrule(lr){3-4} \cmidrule(lr){5-6}
\cellcolor{tabularColor} & \cellcolor{tabularColor} ($\Delta_\text{base}$) & \cellcolor{tabularColor} $\uparrow \Delta H$ & \cellcolor{tabularColor} $\downarrow \Delta H$ & \cellcolor{tabularColor} $\uparrow \Delta H$ & \cellcolor{tabularColor} $\downarrow \Delta H$ (\textbf{Ours}) \\
\midrule
Web. & 13.8 
& \cellcolor{l2} 16.4 \textcolor{up}{($\uparrow 2.6$)} 
& \cellcolor{r1} 11.4 \textcolor{down}{($\downarrow 2.4$)} 
& \cellcolor{r2} 12.4 \textcolor{down}{($\downarrow 1.4$)} 
& \cellcolor{l3} \textbf{18.0} \textcolor{up}{($\mathbf{\uparrow 4.2}$)} \\
HQA & 42.5
& \cellcolor{l1} 44.2 \textcolor{up}{($\uparrow 1.7$)} 
& \cellcolor{r2} 40.5 \textcolor{down}{($\downarrow 2.0$)} 
& \cellcolor{r1} 42.0 \textcolor{down}{($\downarrow 0.5$)} 
& \cellcolor{l2} \textbf{45.8} \textcolor{up}{($\mathbf{\uparrow 3.3}$)} \\
2Wiki. & 40.6
& \cellcolor{l3} 46.8 \textcolor{up}{($\uparrow 6.2$)} 
& \cellcolor{r2} 38.6 \textcolor{down}{($\downarrow 2.0$)} 
& \cellcolor{r1} 39.4 \textcolor{down}{($\downarrow 1.2$)} 
& \cellcolor{l5} \textbf{48.9} \textcolor{up}{($\mathbf{\uparrow 8.3}$)} \\
MuSiQ. & 18.2
& \cellcolor{l1} 19.1 \textcolor{up}{($\uparrow 0.9$)} 
& \cellcolor{r2} 16.4 \textcolor{down}{($\downarrow 1.8$)} 
& \cellcolor{r1} 17.8 \textcolor{down}{($\downarrow 0.4$)} 
& \cellcolor{l2} \textbf{20.5} \textcolor{up}{($\mathbf{\uparrow 2.3}$)} \\
Bamb. & 40.5
& \cellcolor{l1} 41.5 \textcolor{up}{($\uparrow 1.0$)} 
& \cellcolor{r1} 39.8 \textcolor{down}{($\downarrow 0.7$)} 
& \cellcolor{r2} 38.1 \textcolor{down}{($\downarrow 2.4$)} 
& \cellcolor{l3} \textbf{45.9} \textcolor{up}{($\mathbf{\uparrow 5.4}$)} \\
\bottomrule
\end{tabular}
\vspace{-2mm}
\end{table}
\begin{figure}
  \centering
  % \vspace{-4.5mm}
  \includegraphics[width=\linewidth]{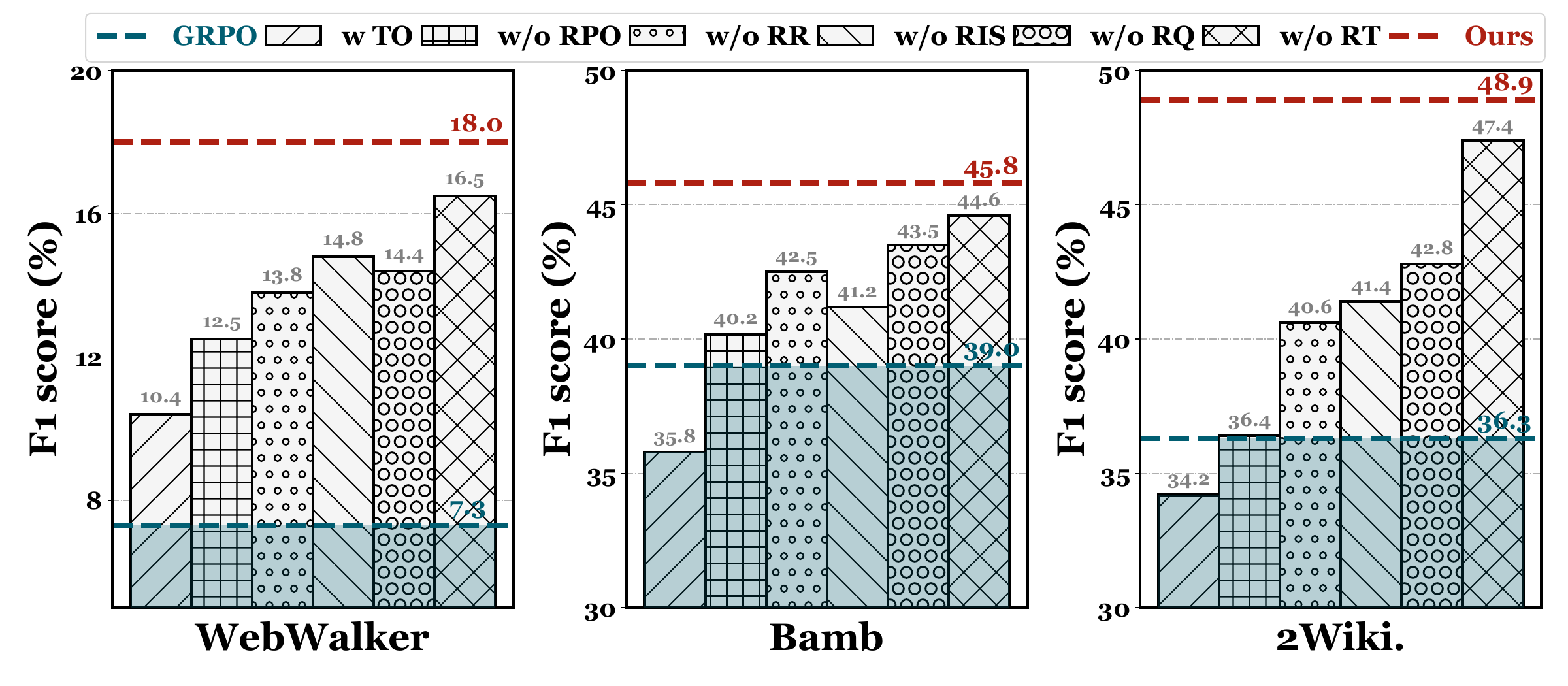} % 0.67
  \vspace{-8mm}
  \caption{Ablation study for model components.}
  % \vspace{-5mm}
  \label{fig:ablation_model}
\end{figure}

\subsection{Ablation Study}\label{sec:ablation}
\subsubsection{Ablation for Model Components}
We begin our ablation study by evaluating the contributions of the key components within {\model}, including the Retrieval-aware Policy Optimization (RPO) in Sec.~\ref{sec:retrieval-aware_policy_optimization}, the Retrieval Reward (RR) in Sec.~\ref{sec:retrieval_reward}, the Retrieval Importance Shaping (RIS) in Sec.~\ref{sec:retrieval_importance_shaping}, the Retrieval Quality (RQ) term in Eq.~\ref{eq:retrieval_reward}, and the Retrieval Timing (RT) term in Eq.~\ref{eq:retrieval_reward}. This results in five variants: \textbf{\textit{w/o RPO}}, \textbf{\textit{w/o RR}}, \textbf{\textit{w/o RIS}}, \textbf{\textit{w/o RQ}}, and \textbf{\textit{w/o RT}}. We also include a Trajectory Off-policy (TO) variant (\textbf{\textit{w/ TO}}), where retrievals are conducted at the trajectory level. The results are shown in Fig.~\ref{fig:ablation_model}. 
We can see that incorporating all components results in the best performance, while the removal of any single component leads to a performance drop. This highlights the effectiveness of each component in {\model}. Notably, the \textit{w/ TO} variant performs worst, underscoring the importance of step-level exploration dynamics during agentic rollout.

\vspace{-1.5mm}
\subsubsection{Ablation for Retrieval Reward}
We further conduct an ablation study to evaluate the designs of retrieval reward, which is formulated to encourage \textbf{entropy reduction} ($\downarrow\Delta H$) at \textbf{high-entropy states} ($H^{\mathrm{High}}$). Specifically, we separately invert this principle via: (i) encouraging \textbf{entropy increase} ($\uparrow\Delta H$) by $H_{\Delta \hat{s}_{t}} \leftarrow -H_{\Delta \hat{s}_{t}}$ in Eq.~\ref{eq:entropy_drop}, and (ii) encouraging retrieval at \textbf{low-entropy states} ($H^{\mathrm{Low}}$) by $H_{\hat{s}_{t-1}} \leftarrow 1/H_{\hat{s}_{t-1}}$ in Eq.~\ref{eq:retrieval_reward}. 
In Tab.~\ref{tab:retrieval_reward}, we find that encouraging entropy reduction at high-entropy states performs best. This may be owing to the effective estimation of retrieval-aware exploration, providing informative signals during policy optimization. Unfortunately, the variants ($H^{\mathrm{Low}}$, $\downarrow\Delta H$) and ($H^{\mathrm{High}}$, $\uparrow\Delta H$) perform poorly (but they still outperform GRPO). We hypothesize that these two settings tend to reinforce the agent's raw reasoning behaviors, preventing it from escaping its native exploration boundary.

\section{Conclusion and Future Work}
In this paper, we investigate Agentic RL and present {\model}, which introduces retrieval to explicitly expand the agent’s step-level exploration capability during training.
By introducing a Hybrid-policy Agentic Rollout strategy, we can enhance rollout diversity using the retrieved off-policy step-level reasoning traces.
Our Retrieval-aware Policy Optimization mechanism then calibrates the policy estimation with retrieval reward and importance shaping, facilitating effective and stable RL training.
As for future work, we will consider more powerful strategies to construct higher-quality Step-Trace Buffers, such as multi-policy frameworks.

\clearpage
\newpage

\bibliographystyle{ACM-Reference-Format}
\bibliography{sample-base}

%%
%% If your work has an appendix, this is the place to put it.

\clearpage
\newpage
\appendix

\vspace{-8mm}
\section{Notations and Algorithms}\label{app_sec:notation}
We provide the important notations used in this paper and their corresponding descriptions as shown in Tab.~\ref{tab:notations}. Additionally, for clarity, we present the pseudo-codes of {\model} in Algorithm~\ref{algorithm}. 

\begin{algorithm}[t]
% \small
  \SetKwInOut{Input}{input}\SetKwInOut{Output}{output}
  \KwIn{Dataset $\mathcal{D}$; LLM agent $\pi_{\theta}$; External tools $\mathcal{T}$; Step-Trace Buffer $\mathcal{B}$; The number of pure on-policy rollouts $N_\mathrm{on}$ and hybrid-policy rollouts~$N_\mathrm{hybrid}$.}
  \For{step = 1 to $S_{\mathrm{total}}$}{
    Initialize old model $\pi_{\theta_{\mathrm{old}}} \leftarrow \pi_{\theta}$ \;
    Sample a batch of training data $q \subseteq \mathcal{D}$ \;
    Initialize the rollout pool $\mathcal{P} \leftarrow \emptyset$ \;
    \For{$i = 1, 2, \cdots, N_{\mathrm{on}} + N_{\mathrm{hybrid}}$}{
        Generate first-step trace $s_0 \sim \pi_{\theta}(\cdot \mid q, t=0)$ \;
        Update rollout pool $\mathcal{P}$ with $s_0$ \;
    }
    Mark $N_{\mathrm{on}}$ rollouts among $\mathcal{P}$ as $\mathcal{P}_\mathrm{on}$ and others as $\mathcal{P}_\mathrm{hybrid}$\;
    \tcp{Hybrid-policy Agentic Rollout}
    \While{$\exists \; \mathcal{S}_{<t} \in \mathcal{P}$ not terminated}{
        \If{$\mathcal{S}_{<t} \in \mathcal{P}_\mathrm{on}$}{
            Generate next trace $s_t \sim \pi_{\theta}(\cdot \mid q, \mathcal{S}_{<t})$ \;
            Update rollout pool $\mathcal{P}$ with $s_t$ \;
        }
        \If{$\mathcal{S}_{<t} \in \mathcal{P}_\mathrm{hybrid}$}{
            $p \sim \mathrm{Random}()$ \;
            \If{$p < 0.5$}{
                Retrieve off-policy trace $\hat{s}_t \sim\mathrm{Retrieve}(\mathcal{S}_{<t})$ \;
                Update rollout pool $\mathcal{P}$ with $\hat{s}_t$ \;
            }
            Generate next trace $s_{t/t+1} \sim \pi_{\theta}(\cdot \mid q, \mathcal{S}_{<t/t+1})$ \;
            Update rollout pool $\mathcal{P}$ with $s_{t/t+1}$ \;
        }
    }
    \tcp{Retrieval-aware Policy Optimization}
    \For{Updating iter = 1, $\cdots$}{
        Compute retrieval reward $Z_\mathrm{ret}$ by Eq.~\ref{eq:retrieval_reward}\;
        Compute retrieval advantage $A_\mathrm{ret}$ by Eq.~\ref{eq:retrieval_advantage}\;
        Compute final advantage $A_\mathrm{RAPO}$ by Eq.~\ref{eq:combined_advantage}\;
        Conduct retrieval importance shaping by Eq.~\ref{eq:retrieval_importance_shaping}\;
        Update the policy model $\pi_\theta$ by Eq.~\ref{eq:final_loss}\;
    }
  }
  \caption{Training LLM agents with {\model}.}
  \label{algorithm}
\end{algorithm}
\vspace{-3mm}

\begin{table}[h]
  \centering
  \caption{Important notations and descriptions.}
  \setlength{\tabcolsep}{1pt}
  \label{tab:notations}
  \small
  \begin{tabular}{cc}
    \toprule
    \cellcolor{tabularColor} \textbf{Notations} & \cellcolor{tabularColor} \textbf{Descriptions}  \\
    \midrule 
    $s_t$  & On-policy reasoning trace at step $t$\\
    $\hat{s}_t$ & Off-policy reasoning trace at step $t$\\
    $\mathcal{S}_{<t}$ & On-policy reasoning history before step $t$\\
    $\hat{\mathcal{S}}_{<t}$ & Off-policy reasoning history before step $t$\\
    \midrule
    $H_{s_t}$ & Step-level entropy for trace $s_t$ \\
    $Z^i_{\mathrm{ret}}$ & Retrieval reward for the $i$-th rollout within group \\
    $\hat{r}_{i,j}(\theta)$ & Retrieval importance shaping for the $j$-th token in $i$-th rollout\\
    $A_{\mathrm{RAPO}}^i$ & Combined advantage for the $i$-th rollout\\
    $J_{\mathrm{RAPO}}(\theta)$ & Training objective of {\model}\\
  \bottomrule
\end{tabular}
\end{table}

\section{Theoretical Analysis}\label{app_sec:theoretical_analysis}
In this section, we conduct a theoretical analysis to validate the effectiveness of {\model}, showing that the Retrieval-aware Policy Optimization mechanism introduces an implicit information bottleneck, which provably enhances generalization in RL training.

\textit{Policy with Retrieval Information.}
We consider an agent whose policy is conditioned on both the environment state and retrieval information. Let us first denote the off-policy-conditioned reasoning process at step $t$ within {\model} as $\pi_\theta(\cdot \mid \hat{\mathcal{S}}_{<t}, t)$, where $\hat{\mathcal{S}}_{<t} = \left\{s_0, \cdots, s_{t-2}, \hat{s}_{t-1}\right\}$ is the reasoning history and $\hat{s}_{t-1} \sim \mathrm{Retrieve}\left(\mathcal{S}_{<t-1}\right)$ is the information from the retrieval process conditioned on the reasoning history at the $(t-1)$-th step. The policy agent is parameterized by $\theta$.

The goal of our Retrieval-aware Policy Optimization mechanism is to train agents that, in addition to maximizing outcome rewards, minimize the impacts and contributions of useless information from the retrieval process. We quantify this using the conditional mutual information $I(\cdot \; ; \; \hat{\mathcal{S}}_{<t} \mid \mathcal{S}_{<t-1})$.

\textit{Information-Regularized Objective.}
This approach of minimizing the impacts and contributions of useless information from the retrieval process can be interpreted as encouraging agents to learn useful reasoning behaviors and to absorb those reasoning behaviors closely, except where diverting from doing so (as a result of 
using information from the retrieval process) leads to higher reward \cite{Distill}.
To see this, the conditional mutual information can be written as:
\begin{equation}
I(\cdot \; ; \; \hat{\mathcal{S}}_{<t} \mid \mathcal{S}_{<t-1})
= \mathbb{E}_{\pi_\theta}
\!\left[
\mathbb{D}_{\mathrm{KL}}\!\left(
\pi_\theta(\cdot \mid \mathcal{S}_{<t-1}, \hat{\mathcal{S}}_{<t})\,\|\,\bar{\pi}_\theta(\cdot \mid \mathcal{S}_{<t-1})
\right)
\right],
\end{equation}
where the expectation is taken over trajectories induced by $\pi_\theta$, and
\[
\bar{\pi}_\theta(\cdot \mid \mathcal{S}_{<t-1})
= \sum_g p(g)\,\pi_\theta(\cdot \mid \mathcal{S}_{<t-1}, g)
\]
is the default policy obtained by marginalizing out retrieval information.
We therefore optimize the following objective:
\begin{equation} \label{eq:info_objective}
\begin{aligned}
J(\theta) & = \mathbb{E}_{\pi_\theta}
\!\left[ 
r - \beta I(\cdot; \hat{\mathcal{S}}_{<t} \mid \mathcal{S}_{<t-1})
\right] \\
& =
\mathbb{E}_{\pi_\theta}
\!\left[
r - \beta
\mathbb{D}_{\mathrm{KL}}\!\left(
\pi_\theta(\cdot \mid \mathcal{S}_{<t-1}, \hat{\mathcal{S}}_{<t})\,\|\,\bar{\pi}_\theta(\cdot \mid \mathcal{S}_{<t-1})
\right)
\right], \\
\end{aligned}
\end{equation}
where $\beta > 0$ regulates the trade-off between reward maximization and information dependence, and $\mathbb{D}_{\mathrm{KL}}(\cdot)$ represents the Kullback-Leibler (KL) divergence.

\textit{Latent Variable Formulation.}
Following prior works \cite{information_1, information_2}, we introduce a latent variable $X$ to parameterize the policy. Specifically, we define an encoder $p_{\mathrm{enc}}(X \mid \mathcal{S}_{<t-1}, \hat{\mathcal{S}}_{<t})$, a decoder $p_{\mathrm{dec}}(\cdot \mid \mathcal{S}_{<t-1}, X)$, and a learned prior $q(X \mid \mathcal{S}_{<t-1})$ such that:
\begin{equation}
    \pi_\theta(\cdot \mid \mathcal{S}_{<t-1}, \hat{\mathcal{S}}_{<t})
= \sum_X p_{\mathrm{enc}}(X \mid \mathcal{S}_{<t-1}, \hat{\mathcal{S}}_{<t})\,p_{\mathrm{dec}}(\cdot \mid \mathcal{S}_{<t-1}, X).
\end{equation}
The latent variable $X$ captures the subset of retrieved information that the agent deems relevant for decision-making at state $\mathcal{S}_{<t-1}$.

\noindent
Under this formulation, the objective in Eq.~\ref{eq:info_objective} admits the following lower bound:
\begin{equation}
\label{eq:lower_bound}
J(\theta)
\ge \tilde{J}(\theta)
=
\mathbb{E}_{\pi_\theta}
\!\left[
r - \beta
\mathbb{D}_{\mathrm{KL}}\!\left(
p_{\mathrm{enc}}(X \mid \mathcal{S}_{<t-1}, \hat{\mathcal{S}}_{<t})\,\|\,q(X \mid \mathcal{S}_{<t-1})
\right)
\right].
\end{equation}

\paragraph{Information Bottleneck.}
According to the data processing inequality (DPI) theory \cite{DPI}, we have:
\[
I(X; \hat{\mathcal{S}}_{<t} \mid \mathcal{S}_{<t-1}) \ge I(\cdot; \hat{\mathcal{S}}_{<t} \mid \mathcal{S}_{<t-1}),
\]
which allows us to upper bound the original information regularizer by bounding
$I(X; \hat{\mathcal{S}}_{<t} \mid \mathcal{S}_{<t-1})$. Conditioning on a fixed state $\mathcal{S}_{<t-1} = s$ and averaging over $p(s)$, we obtain:
\begin{equation}
\label{eq:ib_bound}
I(X; \hat{\mathcal{S}}_{<t} \mid \mathcal{S}_{<t-1})
\le
\sum_s p(s) \sum_g p(g \mid s)\,
\mathbb{D}_{\mathrm{KL}}\!\left(
p(X \mid s, g)\,\|\,r(X)
\right),
\end{equation}
where $r(X)$ is a reference distribution.

This formulation corresponds to an information bottleneck objective, which has been shown to improve generalization and robustness in reinforcement learning \cite{generalization_1, generalization_2}.

\section{Experimental Details}
\subsection{Details of Datasets}\label{app_sec:datasets}
\subsubsection{Datasets of Computational Reasoning}
We first introduce the details of the datasets in Computational Reasoning tasks as follows:
\begin{itemize}[itemsep=3pt, parsep=0pt, leftmargin=*]
\item \textbf{AIME24}\footnote{\url{https://huggingface.co/datasets/HuggingFaceH4/aime_2024}} is designed to assess models’ capabilities in advanced computational reasoning. It contains 30 carefully selected problems drawn from the American Invitational Mathematics Examination. The problems span diverse mathematical domains, including algebraic manipulation and geometric~reasoning. 
% Owing to their high level of difficulty and the diversity of problem formats, AIME24 has become a widely adopted benchmark in recent studies for evaluating reasoning-oriented models.
\item \textbf{AIME25}\footnote{\url{7https://huggingface.co/datasets/math-ai/aime25}} is composed of 30 challenging problems sourced directly from the official AIME I and AIME II examinations released in February 2025. The dataset exhibits broad coverage across fundamental mathematical areas such as algebra, geometry, number theory, and combinatorics. 
% This comprehensive scope makes AIME25 particularly effective at differentiating the computational reasoning proficiency of various models.
\item \textbf{MATH500} \cite{MATH500} is a curated subset of the larger MATH dataset, selected by OpenAI to emphasize more difficult instances. It includes 500 problems that span topics such as algebra, geometry, calculus, and number theory, with difficulty levels approaching or surpassing undergraduate coursework. 
% As a result, it is frequently used in academic evaluations of advanced reasoning models.
\item \textbf{MATH} \cite{MATH} is a widely used benchmark for studying computational reasoning in machine learning models. The dataset covers a broad range of mathematical disciplines, including abstract algebra, calculus, and discrete mathematics. 
% Its training split is organized into multiple difficulty tiers, enabling systematic evaluation of model performance across varying levels of complexity.
\item \textbf{GSM8K} \cite{GSM8K} is a dataset of grade-school–level mathematics problems released by OpenAI. Each problem typically requires between two and eight reasoning steps and involves fundamental arithmetic and logical operations. 
% GSM8K is commonly employed to evaluate models’ step-by-step reasoning abilities and has served as a standard benchmark in numerous studies.
\end{itemize}

\subsubsection{Datasets of Knowledge-Intensive Reasoning}
We then describe the datasets in Knowledge-Intensive Reasoning tasks.
\begin{itemize}[itemsep=3pt, parsep=0pt, leftmargin=*]
\item \textbf{HotPotQA} \cite{HotpotQA} is a benchmark dataset for multi-hop question answering, where all supporting documents are drawn from Wikipedia. Benefiting from Wikipedia’s broad coverage and well-organized content, the dataset provides a rich knowledge source for complex reasoning tasks. 
% As a result, HotPotQA is widely used to evaluate the ability of large language models to handle multi-step information retrieval and reasoning.
\item \textbf{2WikiMultihopQA} \cite{2wiki} is specifically constructed to assess multi-hop question answering. The dataset focuses on evaluating whether models can perform step-by-step reasoning while aggregating evidence from multiple documents, placing strong emphasis on cross-document information integration.
\item \textbf{MuSiQue} \cite{Musique} is a dataset for multi-hop question answering, which is designed to move beyond shallow fact lookup and instead measure a model’s capacity for deeper semantic understanding and logical reasoning across multiple pieces of evidence.
\item \textbf{Bamboogle} \cite{Bamboogle} is a multi-hop question-answering dataset. Its test split is relatively small, containing only 125 question–answer pairs, but it is often used to closely examine model behavior in controlled multi-hop reasoning scenarios.
\end{itemize}

\subsubsection{Datasets of Web-Agentic Reasoning}
Finally, we introduce the datasets in Web-Agentic Reasoning tasks.
\begin{itemize}[itemsep=3pt, parsep=0pt, leftmargin=*]
\item \textbf{SimpleQA} \cite{Bamboogle} is a factual question–answering benchmark consisting of 500 short, knowledge-seeking QA pairs. The dataset is adversarially constructed with respect to GPT-4, meaning that the questions are intentionally selected to expose cases where strong language models are prone to making confident but incorrect factual claims.
\item \textbf{GAIA} \cite{GAIA} is a comprehensive benchmark designed to evaluate the capabilities of general-purpose AI assistants on real-world tasks. The questions in GAIA require a diverse set of skills, including multi-step reasoning, tool usage, web interaction, and multi-modal understanding.
\item \textbf{WebWalkerQA} \cite{webWalker} is a dataset aimed at assessing an agent’s ability to perform structured web navigation in order to answer questions. It contains 680 question–answer tasks that require sequential web traversal and information gathering. 
% The dataset is organized into three difficulty tiers, reflecting increasing complexity in navigation depth and reasoning requirements.
\item \textbf{BrowseComp} \cite{Browsecomp} is a challenging benchmark for evaluating agents’ proficiency in web browsing and complex information discovery. The dataset consists of 1,266 questions that involve obscure, fragmented, or difficult-to-locate information on the~web. 
% Successfully answering these questions often requires iterative exploration, careful evidence aggregation, and robust reasoning.
\end{itemize}

\subsubsection{Datasets for RL Training}
\begin{itemize}[itemsep=3pt, parsep=0pt, leftmargin=*]
\item \textbf{Computational and Knowledge-Intensive Reasoning.} We adopt the training dataset from Tool-Star \cite{ToolStar} due to its higher quality and widespread adoption. It includes about 10k questions from both computational reasoning and multi-hop QA reasoning.
\item \textbf{Web-Agentic Reasoning.} We follow the training setup of Tree-GRPO \cite{ToolStar}, which is lightweight and cost-efficient for real-world web API usage. The dataset includes 2k samples from ASearcher-35K \cite{ASearcher-35K} and 200 hard web QA pairs from WebDancer~\cite{WebDancer}.
\end{itemize}

\subsection{Details of Baselines}\label{app_sec:baselines}
\subsubsection{Tool-Integrated Reasoning Methods}
We include three recent tool-integrated reasoning methods for comparison.
\begin{itemize}[itemsep=3pt, parsep=0pt, leftmargin=*]
\item \textbf{Search-o1} \cite{Search-o1} is a RAG-style approach centered on a reasoning model. When search is triggered, the model formulates search queries, retrieves evidence, and then consolidates the returned content into an intermediate context that is woven into the ongoing reasoning trace. The model subsequently continues generation conditioned on this augmented reasoning context until producing the final answer.
\item \textbf{Search-R1} \cite{Search-r1} is an RL-based method that learns a search-calling policy from scratch. Conceptually, its inference workflow resembles Search-o1: the model emits queries as needed during multi-step reasoning, fetches relevant documents from a search engine, and integrates the retrieved evidence back into its reasoning trajectory before continuing.
\item \textbf{Tool-Star} \cite{ToolStar} is a tool-augmented reasoning framework that supports six different tool types and emphasizes systematic design in both data construction and model training. It introduces an automated data synthesis pipeline that generates tool-invocation trajectories by combining tool-aware prompting with hint-guided sampling, which are further refined through quality normalization and difficulty-based filtering.
\end{itemize}

\subsubsection{Off-policy Learning Methods}
We also include the existing off-policy learning methods as follows:
\begin{itemize}[itemsep=3pt, parsep=0pt, leftmargin=*]
\item \textbf{RolloutReplay} \cite{ReplayBuffer} aims to improve training efficiency of RL for single-step reasoning. It incorporates a rollout replay mechanism inspired by experience replay in classical reinforcement learning, allowing recent rollouts to be reused to reduce computational overhead while preserving training stability.
\item \textbf{LUFFY} \cite{LUFFY} is designed to overcome the on-policy limitation for single-step reasoning. It jointly leverages off-policy trajectory and on-policy rollouts, which builds on a mixed-policy GRPO formulation with theoretical convergence guarantees and further applies policy shaping through regularized importance sampling.
\end{itemize}

\subsubsection{RL Methods}
Now, we introduce the details of the RL methods used in this paper.
\begin{itemize}[itemsep=3pt, parsep=0pt, leftmargin=*]
\item \textbf{GRPO} \cite{GRPO} is a single-step RL approach for LLM fine-tuning that operates through group-level policy optimization. It assesses multiple sampled outputs jointly and derives relative reward signals from within-group comparisons, leading to more stable updates and improved sample efficiency during training.
\item \textbf{DAPO} \cite{DAPO} improves optimization stability by separating the clipping mechanism from the policy update process. It employs an adaptive data selection strategy that dynamically chooses training samples to sustain informative gradients into learning.
\item \textbf{GPPO} \cite{GPPO} modifies the standard PPO objective by disentangling the clipping operation across forward and backward computations. Specifically, clipping is applied during the forward pass to constrain policy updates, while the backward pass retains the unclipped policy ratio to preserve richer gradient information.
\item \textbf{GiGPO} \cite{GiGPO} is a multi-step RL that computes advantages at multiple granularities. At the episode level, entire trajectories are grouped to estimate coarse-grained advantage signals; At the step level, actions associated with shared anchor states are regrouped across trajectories to derive fine-grained credit assignments. 
\item \textbf{Tree-GRPO} \cite{Tree-GRPO} is a multi-step RL that leverages tree-structured search. It represents each agent interaction step as a node in a search tree, allowing multiple rollouts to share common prefixes. Furthermore, the tree-based trajectories enable the extraction of step-level supervision signals. 
\item \textbf{ARPO} \cite{ARPO} is designed for multi-step LLM agents operating in interactive environments. It leverages an entropy-driven adaptive rollout strategy that increases exploration in high-uncertainty steps and incorporates a structured advantage attribution mechanism to distribute credit across branching paths.
\item \textbf{AEPO} \cite{AEPO} is a multi-step RL designed to address entropy-related challenges during both rollout generation and policy optimization. It introduces an entropy-aware rollout strategy and an entropy-balanced policy optimization objective to mitigate excessive branching and ensure stable gradient scaling. 
\end{itemize}

\vspace{-3mm}
\subsection{Details of Experimental Settings}\label{app_sec:settings}
\subsubsection{Buffer Construction}
During the Step-Trace Buffer construction, we use AEPO-Qwen3-14B \cite{AEPO} as the default off-policy agent. We set $N=16$ during the generation of off-policy trajectories and set $K = 5$ in reward-aware filtering for each query within the training dataset. The scale of our Step-Trace Buffer used in each RL training is provided in the following subsections.
The resulting reasoning corpus can be consistently reused, and we will make it publicly available if this paper can be accepted.

\subsubsection{RL Training}
We build upon the Search-R1 \cite{Search-r1} pipeline implemented in the VeRL \cite{verl} framework.

For Computational Reasoning and Knowledge-Intensive Reasoning, we use a total training batch size of 32, a PPO mini-batch size of 16, and a global rollout size of 16 across all experiments.
For {\model}, we set the number of hybrid-policy trajectories $N_\mathrm{hybrid}$ to 8, the importance shaping weight $m$ to 0.05, and the retrieval advantage weight $a$ to 0.2.
The KL divergence coefficient and clipping parameter are set to 0 and 0.28, respectively.
The constructed buffer in this setting contains 50,000 off-policy trajectories, comprising 169,489 step-level reasoning traces with a total of 15,648,438 tokens.

For Web-Agentic Reasoning, we adopt the configuration of Tree-GRPO \cite{Tree-GRPO} and set the training batch size of 128, a PPO mini-batch size of 64, and a global rollout size of 4 across this group of experiments.
During each search call, the top-10 snippets derived from the web SERP API are provided to the agent as external evidence.
For {\model}, we set $N_{\mathrm{hybrid}}$ to 2 while keeping other settings consistent with those in Computational and Knowledge-Intensive Reasoning tasks.
The resulting buffer contains 11,000 off-policy trajectories, comprising 38,473 step-level reasoning traces and 4,518,194 tokens in total.
Unless specified, experiments are performed on 8 NVIDIA A100 GPUs. Off-policy retrieval is not used during~evaluation.

\subsubsection{Details of Search Tool}
For Computational and Knowledge-Intensive Reasoning, we employ a widely used Wikipedia dump \cite{Wikipedia} as the corpus and adopt MiniLM \cite{MiniLM} as the semantic encoder for document embeddings. During training, we follow the Search-R1 \cite{Search-r1} protocol and apply the RAG pipeline to derive the top-3 most semantically relevant documents for each search step. We also use this pipeline to perform the retrieval within the hybrid-policy rollout of {\model}, which returns the most relevant off-policy trace.

For Web-Agentic Reasoning, we use the Bing Search API (US-EN region) as the search engine. At each search step, the top-10 passages are provided to the agent as search feedback.

\begin{figure*}
  \centering
  \begin{subfigure}{0.3\linewidth}
    \includegraphics[width=.8\linewidth]{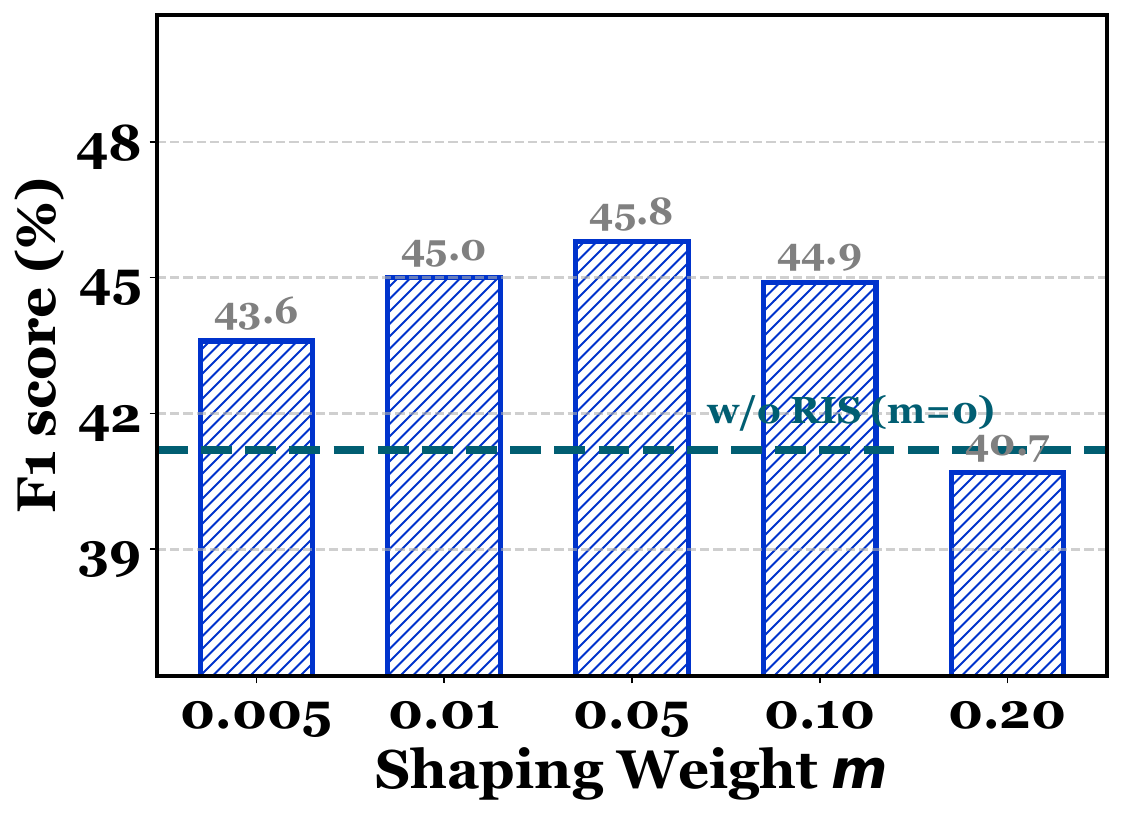}
    \vspace{-2mm}
    \caption{Retrieval importance shaping weight $m$.}
  \end{subfigure}
  \begin{subfigure}{0.3\linewidth}
    \includegraphics[width=.8\linewidth]{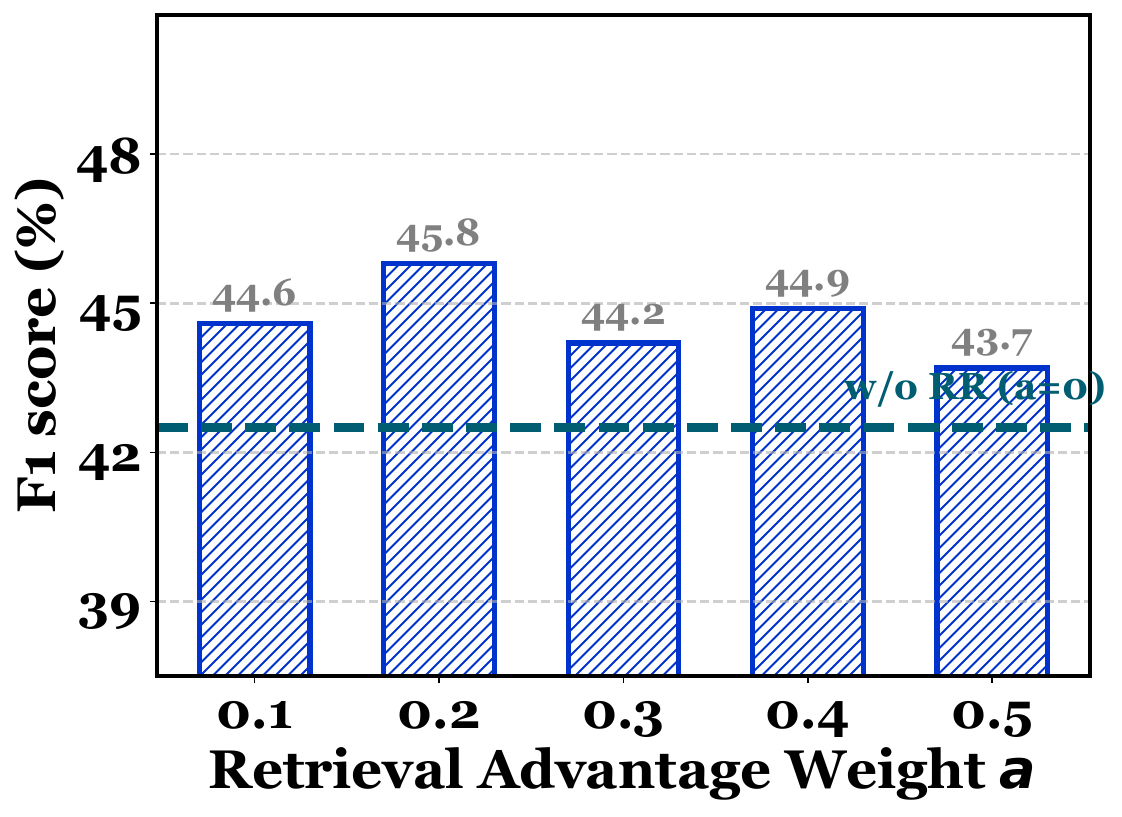}
    \vspace{-2mm}
    \caption{Retrieval advantage weight $a$.}
  \end{subfigure}
  \begin{subfigure}{0.3\linewidth}
    \includegraphics[width=.8\linewidth]{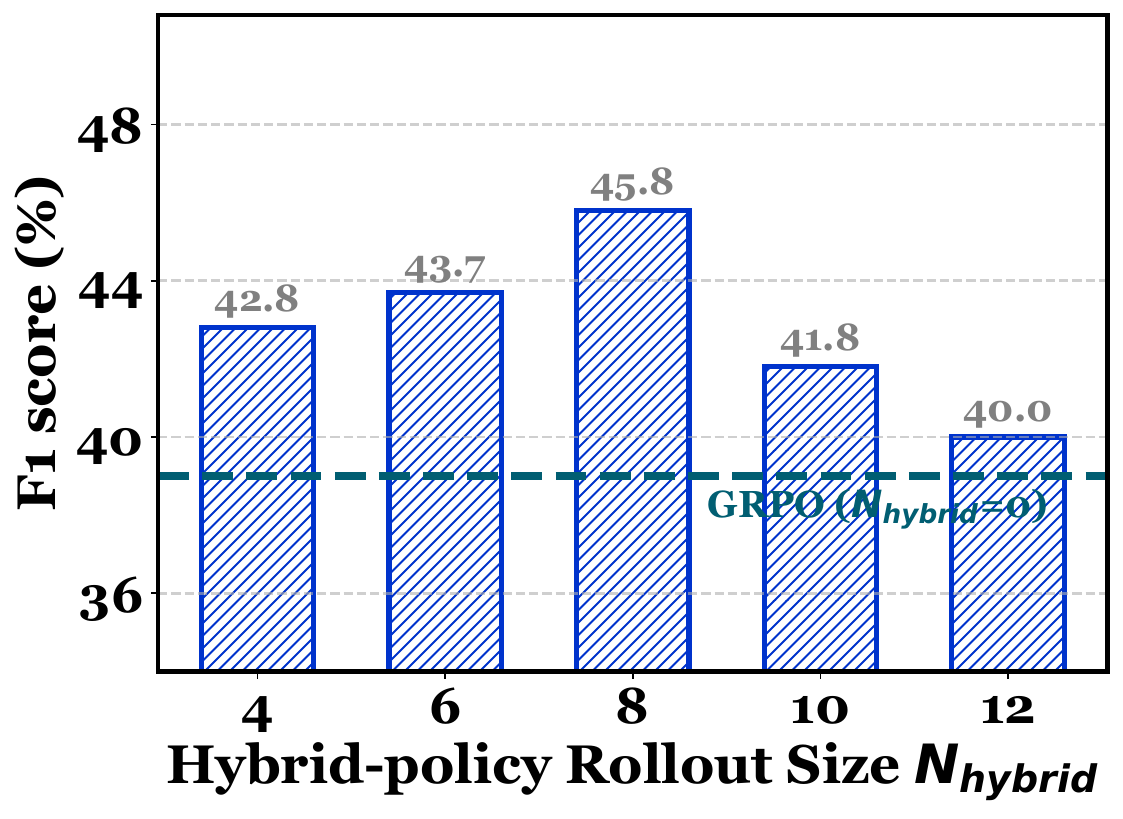}
    \vspace{-2mm}
    \caption{Hybrid-policy rollout size $N_\mathrm{hybrid}$.}
  \end{subfigure}
  \vspace{-2mm}
  \caption{Parameter study on the HQA dataset.}
  \vspace{-2mm}
  \label{app_fig:para}
\end{figure*}

% \vspace{-10mm}
\section{Prompt Template}
In this section, we describe the prompt template used during {\model} training. Following ARPO \cite{ARPO}, any content enclosed within the \texttt{<search>} \texttt{</search>} or \texttt{<python>} \texttt{</python>} tags is parsed as a tool invocation, corresponding to the agent’s action $\alpha$. The tool outputs are subsequently wrapped within \texttt{<result>} \texttt{</result>} tags and returned to the agent as observations $o$, thereby forming a complete one-step reasoning trace $s_t = (\tau_t, \alpha_t, o_t)$.

Different from existing Agentic RL methods, {\model} explicitly exposes retrieved off-policy step traces to the agent. Specifically, the retrieved off-policy traces are inserted into the prompt and delimited by \texttt{<retrieve>} \texttt{</retrieve>} tags, allowing the agent to condition its subsequent reasoning on externally retrieved behaviors in a structured and transparent manner. The overall prompt template used in {\model} is summarized as follows:

\vspace{-2mm}
\begin{templatebox}{Prompt Template for {\model} Training}
\vspace{-1mm}
\textbf{system}

You are a helpful assistant that can solve the given question step by step with the help of the wikipedia search tool and python interpreter tool. Given a question, you need to first think about the reasoning process in the mind and then provide the answer. During thinking, you can invoke the wikipedia search tool to search and python interpreter tool to calculate the math problem for fact information about specific topics if needed. The reasoning process and answer are enclosed within <think> </think> and <answer> </answer> tags respectively, and the search query and result are enclosed within <search> </search> and <result> </result> tags respectively. 
\textit{Additional context may be enclosed in <retrieve> </retrieve> tags. Use it if helpful.}
For example, <think> This is the reasoning process. </think> <search> search query here </search> <result> search result here </result> <think> This is the reasoning process. </think> <python> python code here </python> <result> python interpreter result here </result> <think> This is the reasoning process. </think> <answer> The final answer is [ \string\boxed\{answer here\} ] </answer>. In the last part of the answer, the final exact answer is enclosed within \string\boxed\{\} with latex format.

\textbf{user}

Euxoamorpha eschata is a moth found in a city situated on the southern shores of what?

\textbf{assistant}
\end{templatebox}

\begin{figure}
  \centering
  %--- A 独占一行 ---
  \begin{subfigure}{0.5\linewidth}
  \hspace{-5mm}
    \includegraphics[width=1.1\linewidth]{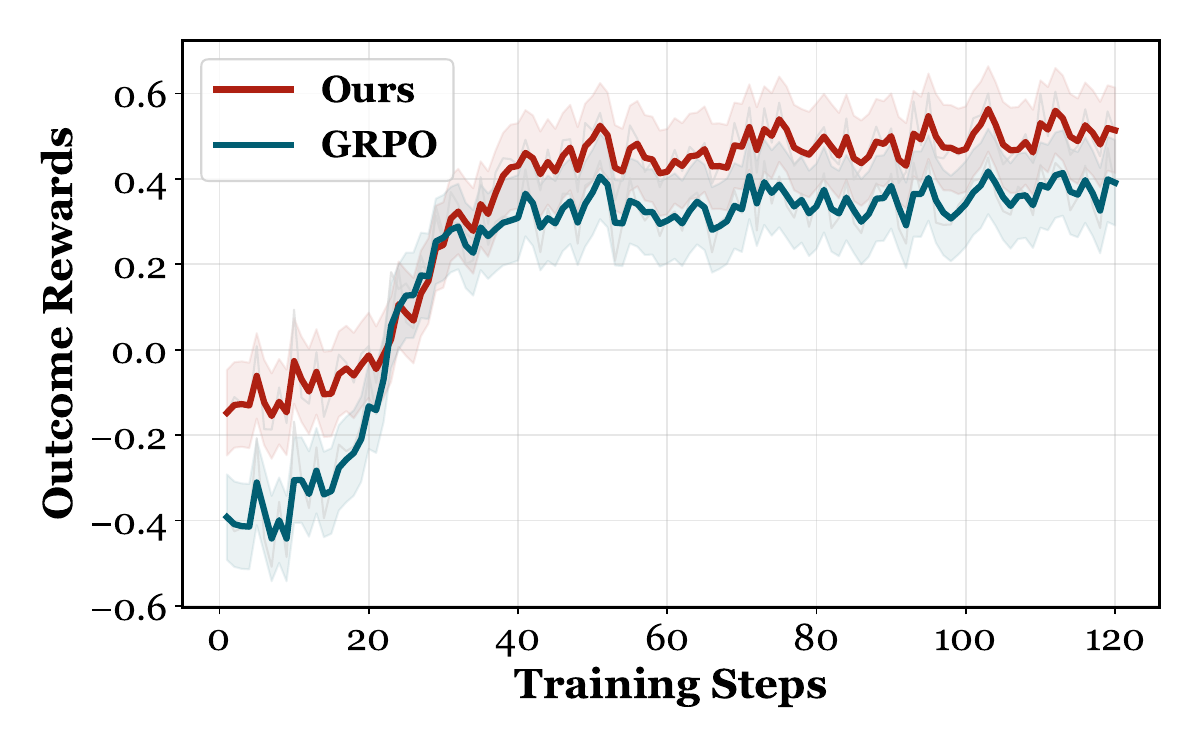}
    \caption{Outcome rewards.}
    \label{PlaceHolder}
  \end{subfigure}
  \hspace{-2mm}
  \begin{subfigure}{0.5\linewidth}
    \includegraphics[width=1.1\linewidth]{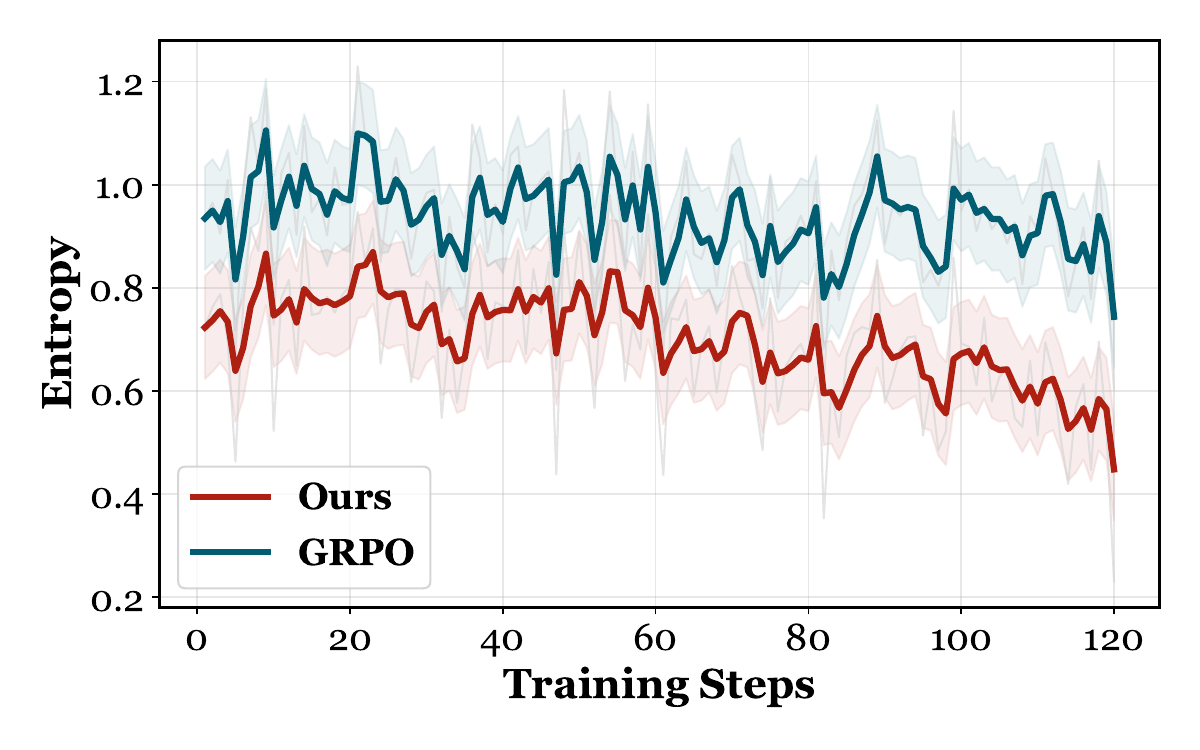}
    \caption{Entropy.}
    \label{fig:PlaceHolder}
  \end{subfigure}
  \caption{Training dynamics between {\model} and GRPO.}
  \label{app_fig:training_dynamics}
\end{figure}
\begin{figure}
  \centering
  %--- A 独占一行 ---
  \begin{subfigure}{0.5\linewidth}
  \hspace{-5mm}
    \includegraphics[width=1.1\linewidth]{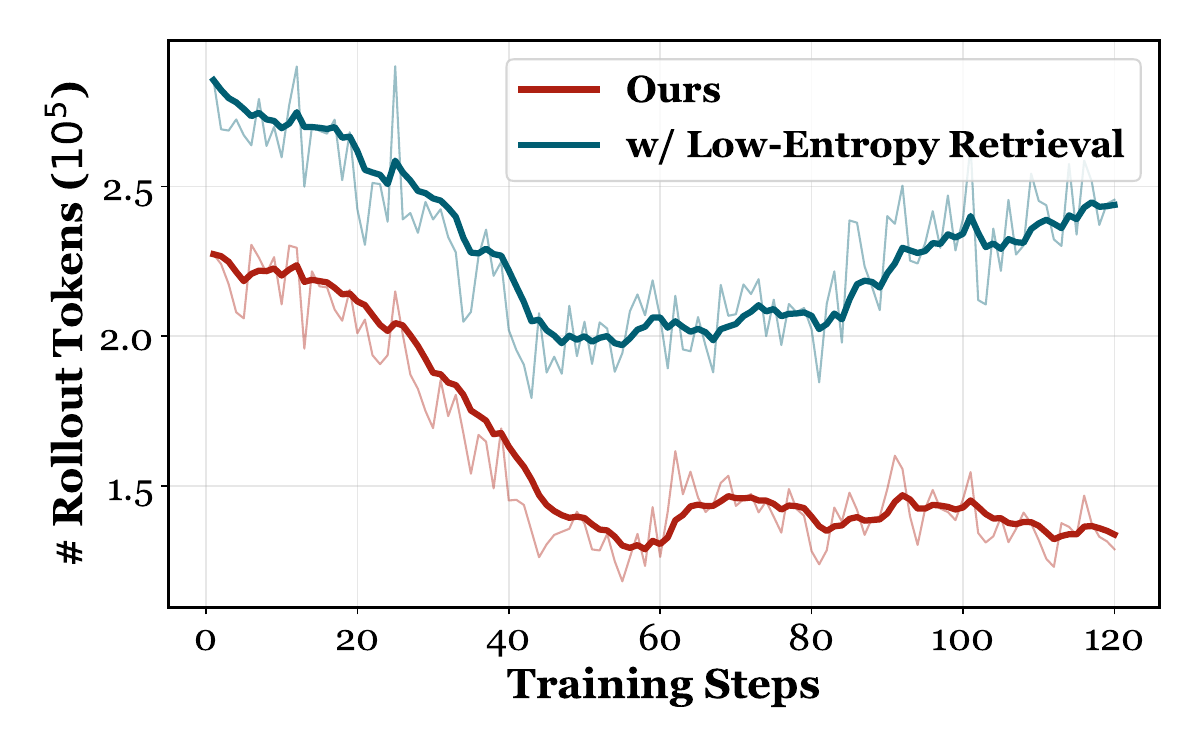}
    \caption{Rollout tokens.}
    \label{PlaceHolder}
  \end{subfigure}
  \hspace{-2mm}
  \begin{subfigure}{0.5\linewidth}
    \includegraphics[width=1.1\linewidth]{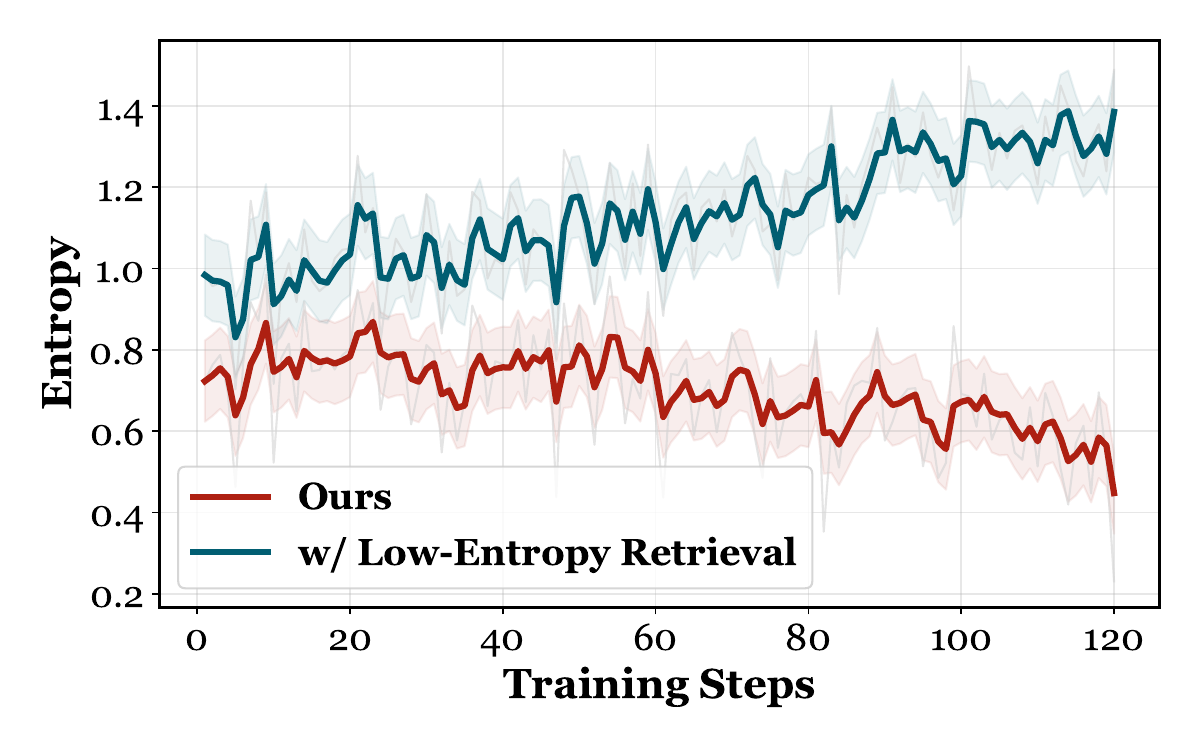}
    \caption{Entropy.}
    \label{fig:PlaceHolder}
  \end{subfigure}
  \caption{Visualization for the \textit{w/ Low-Entropy Retrieval} variant. Encouraging retrieval at low-entropy states tends to force the agent into excessive overthinking during rollout.}
  \label{app_fig:low-entropy-retrieval}
\end{figure}

\begin{figure}
  \centering
  %--- A 独占一行 ---
  \begin{subfigure}{0.5\linewidth}
  \hspace{-5mm}
    \includegraphics[width=1.1\linewidth]{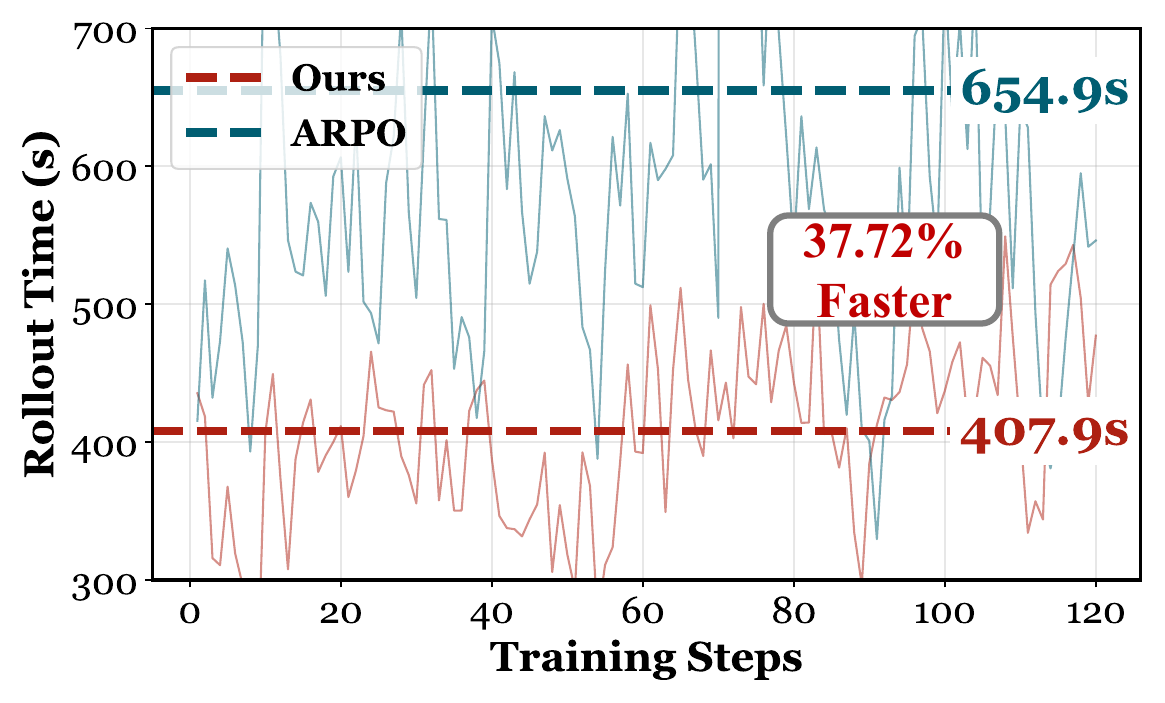}
    \caption{Rollout time.}
    \label{PlaceHolder}
  \end{subfigure}
  \hspace{-2mm}
  \begin{subfigure}{0.5\linewidth}
    \includegraphics[width=1.1\linewidth]{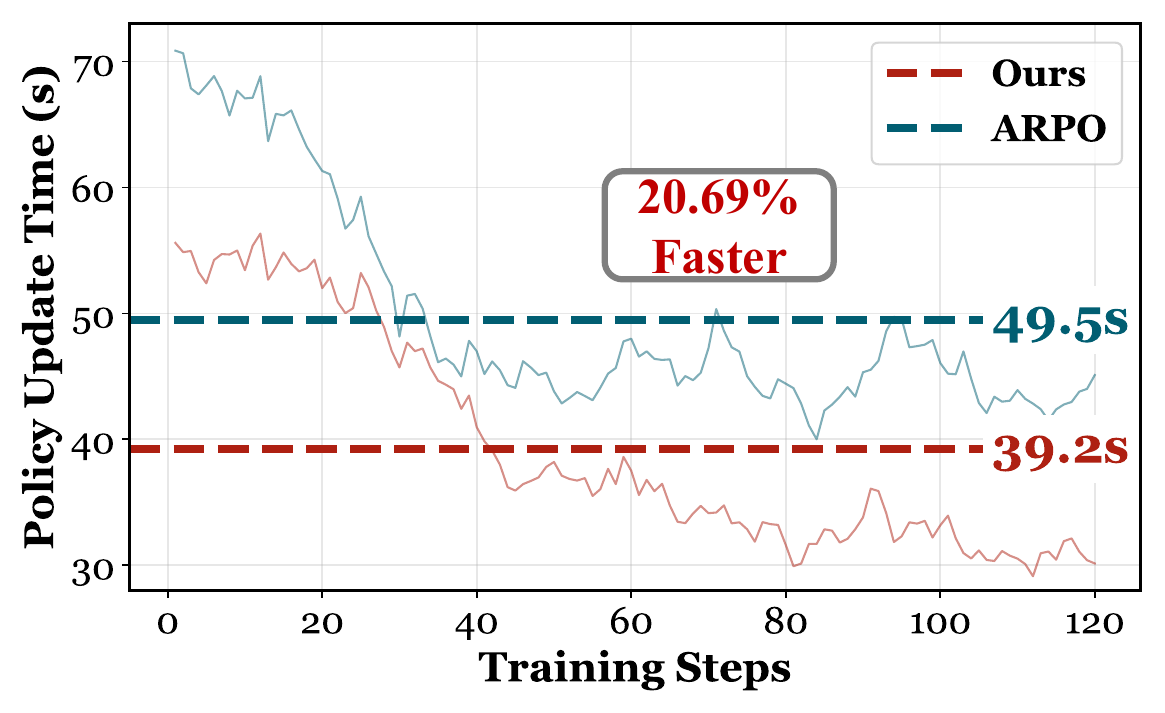}
    \caption{Policy update time.}
    \label{fig:PlaceHolder}
  \end{subfigure}
  \begin{subfigure}{0.5\linewidth}
  \hspace{-5mm}
    \includegraphics[width=1.1\linewidth]{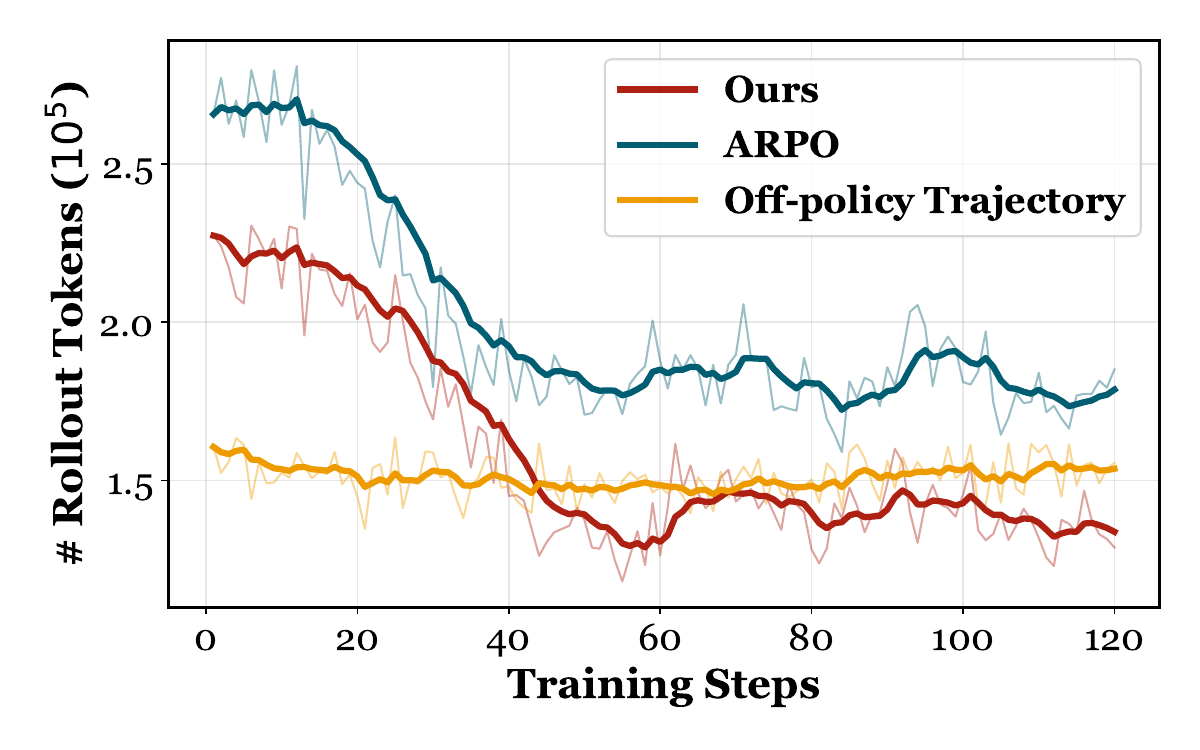}
    \caption{Rollout tokens.}
    \label{PlaceHolder}
  \end{subfigure}
  \hspace{-2mm}
  \begin{subfigure}{0.5\linewidth}
    \includegraphics[width=1.1\linewidth]{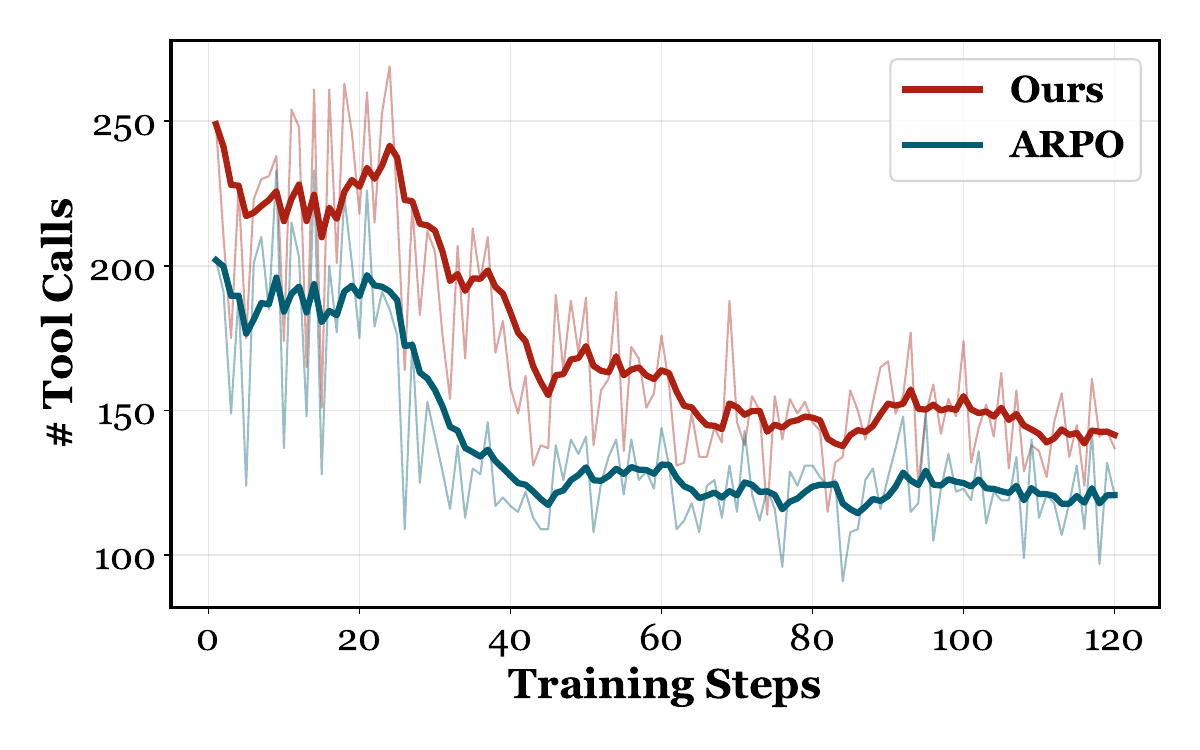}
    \caption{Tool calls.}
    \label{app_fig:PlaceHolder}
  \end{subfigure}
  \caption{Efficiency study over ARPO \cite{ARPO}. {\model} exhibits substantial efficiency gains in rollout time, policy update time, and the number of rollout tokens. {\model} incurs slightly more tool calls than ARPO. This is expected as ARPO reduces up to 50\% of tool calls via branching, whereas {\model} mitigates tool usage via retrieval, avoiding about 25\% of tool calls.}
  \label{app_fig:effiency}
\end{figure}

\section{Training Dynamics}
In this section, we analyze the training dynamics of {\model}, as illustrated in Fig.~\ref{app_fig:training_dynamics}. We report the averaged outcome rewards and the entropy during training. Overall, {\model} exhibits consistently higher outcome rewards than GRPO. Meanwhile, its entropy decreases more steadily, suggesting that the agent learns to produce more confident and coherent reasoning behaviors. These findings highlight the role of retrieval-augmented exploration in improving training effectiveness.

\section{Parameter Study}
Now, we study how the hyper-parameters impact the performance of {\model}. Specifically, we vary: (i) the importance sampling weight $m$ in Eq.~\ref{eq:retrieval_importance_shaping}; (ii) the retrieval advantage weight $a$ in Eq.~\ref{eq:combined_advantage}; and (iii) hybrid-policy rollout size $N_\mathrm{hybrid}$ in Sec.~\ref{sec:hybrid_policy_agentic_rollout}. Note that we fix the global rollout size to 16 across all experiments.

We plot the results in Fig.~\ref{app_fig:para}. Over-emphasizing retrieval-related signals (large $m$, $a$, or $N_{\mathrm{hybrid}}$) introduces excessive off-policy noise that impairs the agent's on-policy self-reasoning, thus leading to degraded performance. Conversely, under-utilizing retrieval signals (small $m$, $a$, or $N_{\mathrm{hybrid}}$) prevents the agent from effectively absorbing and exploiting external reasoning behaviors, limiting the benefits of retrieval-aware exploration.
Overall, setting $m=0.05$, $a=0.2$, and $N_{\mathrm{hybrid}}=8$ seems to be a generally sweet choice.

\section{What Happens When Encouraging Retrieval at Low-Entropy States?}
To further validate the design rationale of our retrieval reward, we conduct a focused analysis on the variant \textit{w/ Low-Entropy Retrieval}. In this setting, retrieval is explicitly encouraged at low-entropy states by modifying the timing term in Eq.~\ref{eq:retrieval_reward} as $H_{\hat{s}_{t-1}} \leftarrow 1 / H_{\hat{s}_{t-1}}$, consistent with the ablation configuration in Sec.~\ref{sec:ablation}. 
We track the number of rollout tokens and entropy respectively and  summarize the results in Fig.~\ref{app_fig:low-entropy-retrieval}. Compared to {\model}, \textit{w/ Low-Entropy Retrieval} exhibits a clear increasing trend in both rollout token length and entropy as training progresses. This indicates that encouraging retrieval at low-uncertainty states forces the agent to excessively \textit{overthink} during rollout, which ultimately degrades reasoning quality and leads to inferior performance (See Tab.~\ref{tab:retrieval_reward}). 

\begin{figure}
  \centering
  \begin{subfigure}{0.5\linewidth}
    \includegraphics[width=1.01\linewidth]{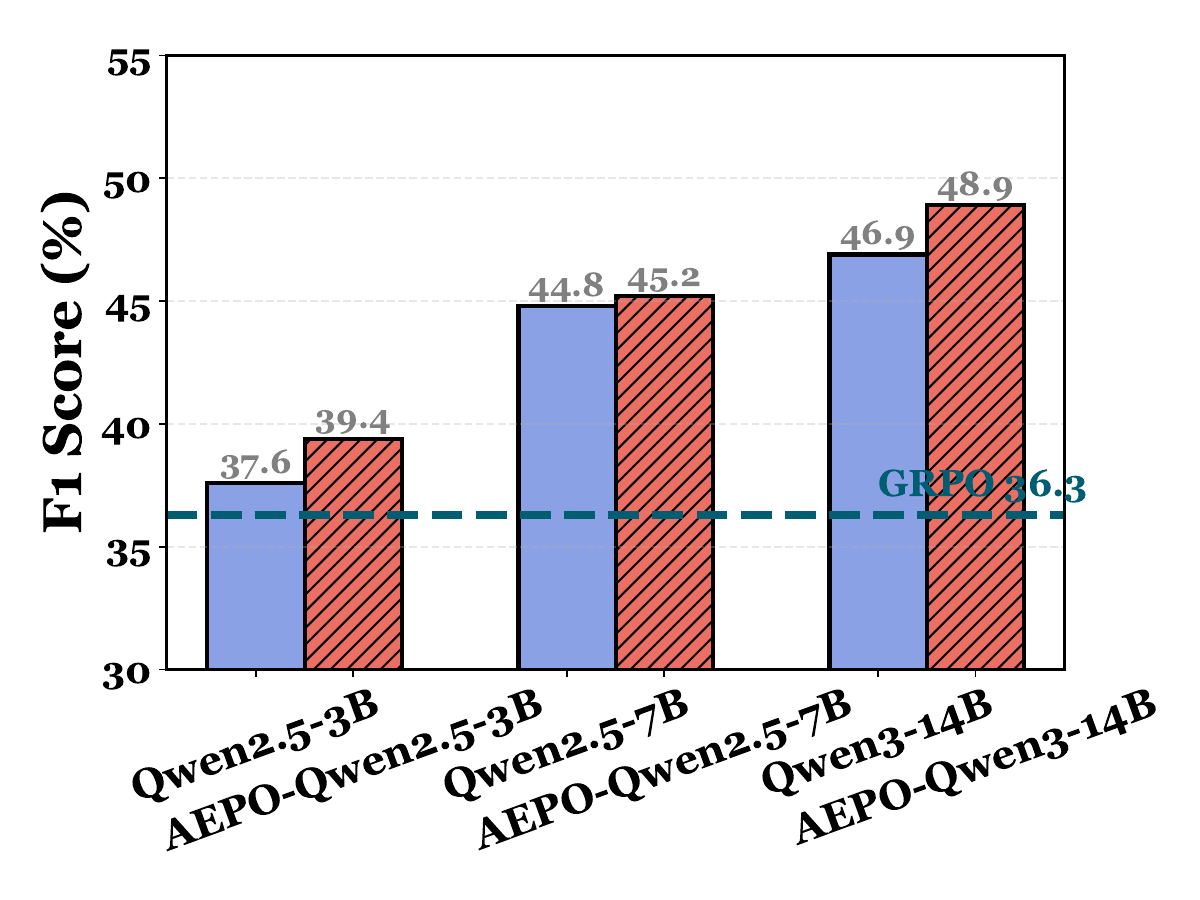}
    \caption{2Wiki.}
  \end{subfigure}
  \hspace{-2mm}
  \begin{subfigure}{0.5\linewidth}
    \includegraphics[width=1.01\linewidth]{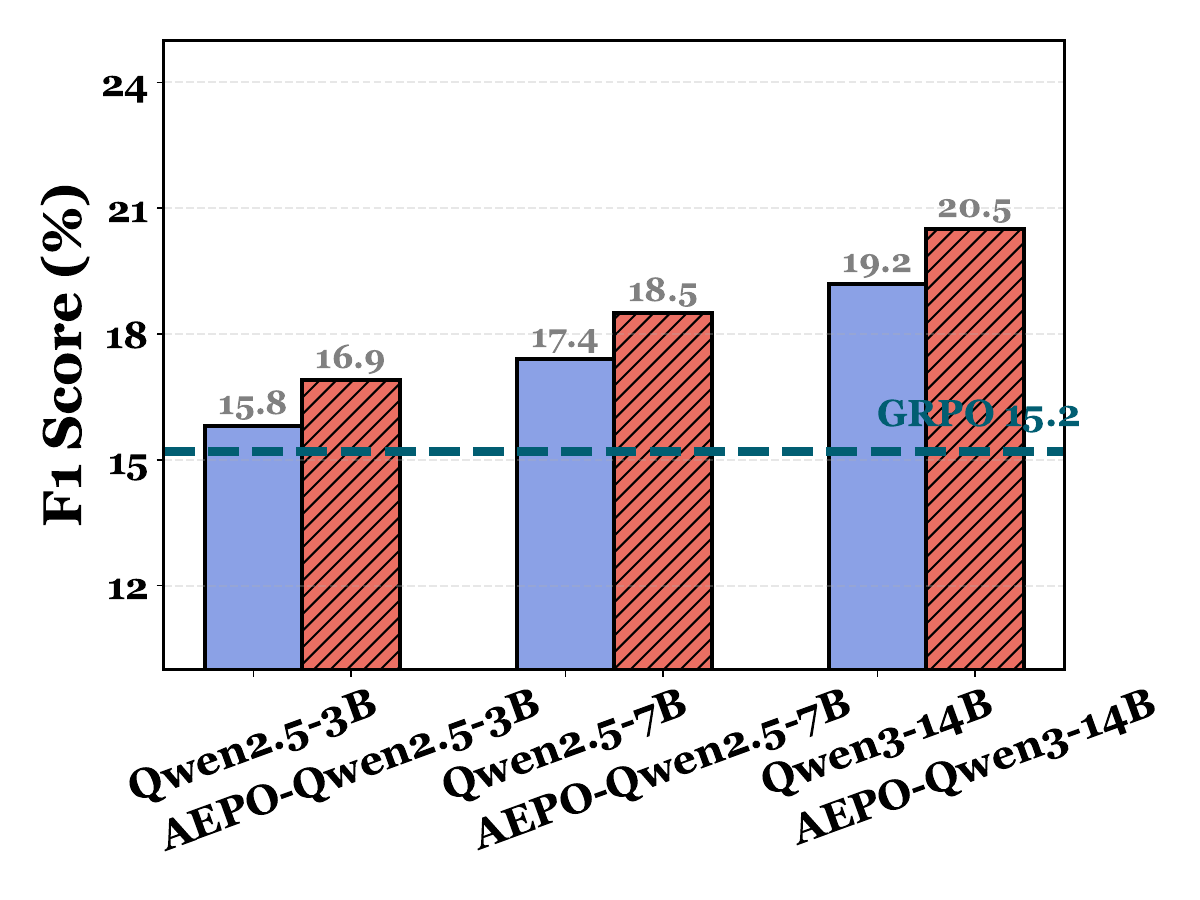}
    \caption{MuSiQ.}
  \end{subfigure}
  \caption{Robustness study for different off-policy models.}
  \label{app_fig:robust_off_policy_model}
\end{figure}
\begin{figure}
  \centering
  \begin{subfigure}{0.5\linewidth}
    \includegraphics[width=1.01\linewidth]{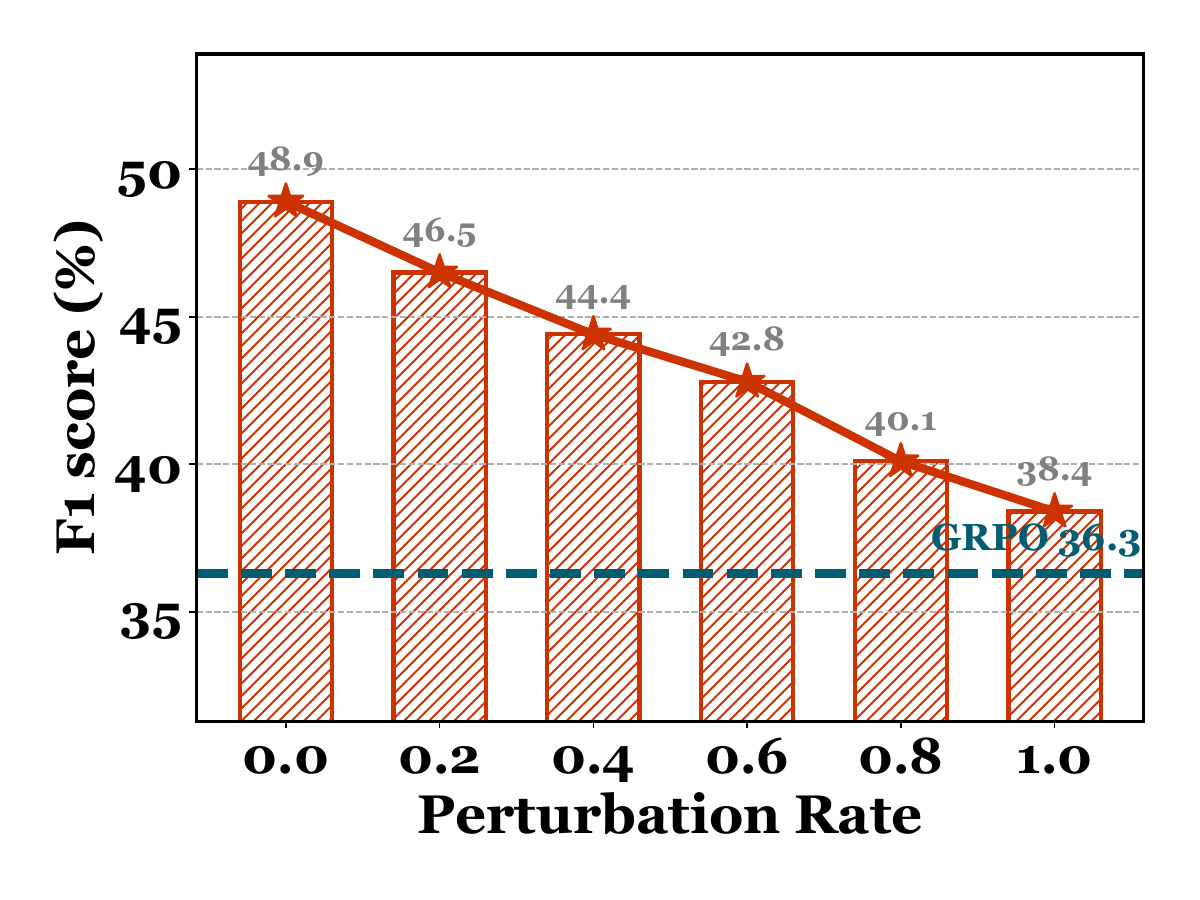}
    \caption{2Wiki.}
  \end{subfigure}
  \hspace{-2mm}
  \begin{subfigure}{0.5\linewidth}
    \includegraphics[width=1.01\linewidth]{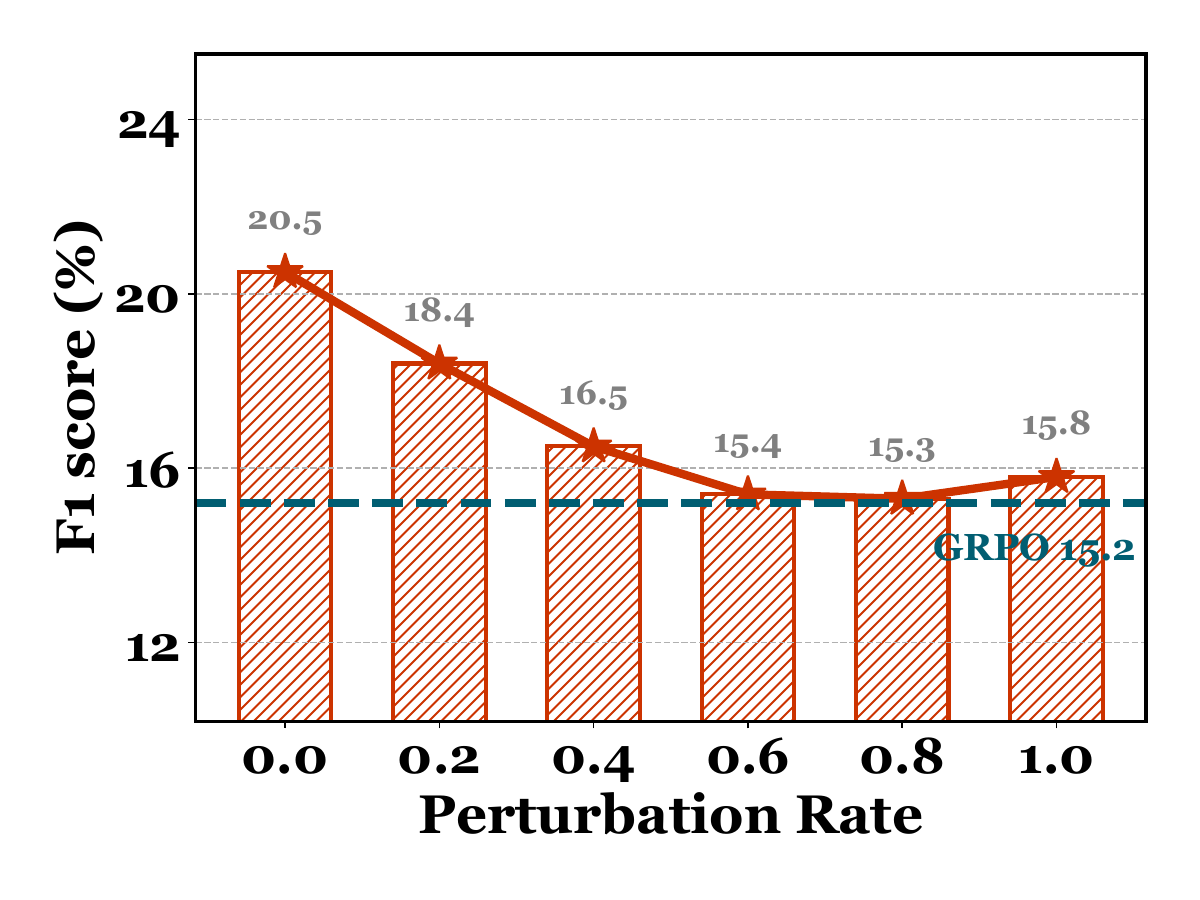}
    \caption{MuSiQ.}
  \end{subfigure}
  \caption{Robustness study for noisy retrieval.}
  \label{app_fig:robust_noise}
\end{figure}

\begin{figure}
  \centering
  \includegraphics[width=\linewidth]{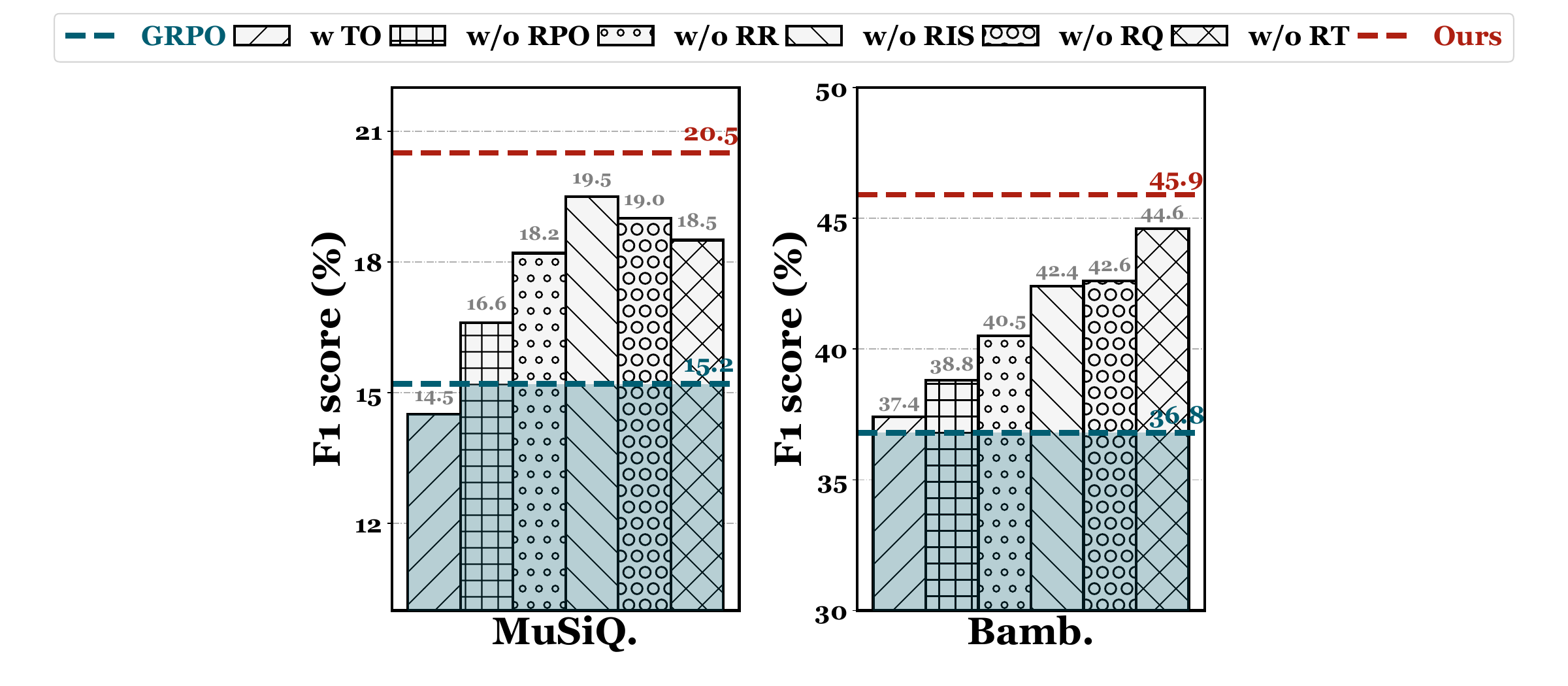} % 0.67
  \caption{Ablation study for model components.}
  \label{app_fig:ablation_model}
\end{figure}

% \section{Failure Attempts}

\section{Limitation}
One potential limitation of {\model} is that, while {\model} significantly expands the on-policy exploration space, the exploration is still implicitly bounded by the joint capacity of the on-policy agent and the off-policy models. 
Future research could investigate more generalized buffer construction paradigms, such as incorporating multiple heterogeneous off-policy agents or dynamically evolving off-policy models, to further broaden exploration in Agentic RL.

Another limitation is that {\model} needs to maintain a Step-Trace Buffer that stores off-policy reasoning traces for retrieval. This introduces additional memory and storage overhead during training, particularly in large-scale post-training. While our experiments demonstrate that retrieval-aware exploration yields substantial performance gains, future work could explore more efficient and scalable strategies for buffer construction.

\section{Case Study}
In this section, we present a qualitative Case Study for the models trained with {\model}. As shown in Tabs.~\ref{tab:case_study_1}-\ref{tab:case_study_4}, the model successfully executes multi-turn agentic tasks through iterative tool invocations and autonomous information aggregation. Interestingly, the model exhibits reflective behaviors during reasoning, suggesting that it has effectively internalized beneficial external behaviors and developed more structured and effective reasoning patterns. These observations further demonstrate the effectiveness of {\model}.

\begin{table*}[h]
    \caption{
        Case Study (i).
    }
    \label{tab:case_study_1}
    \begin{tabularx}{\linewidth}{X}
        \toprule
        \rowcolor{gray!20}
        \textbf{Example from HQA dataset} \\ 
        \midrule
        \textbf{Question}: What British singer-songwriter hosted the 16th Annual Young Hollywood Awards? \\
        \midrule
        \textbf{Ground Truth}: Kelly Lee Osbourne \\
        \midrule
        \rowcolor{gray!20}
        \textbf{Model Output by RAPO-Qwen2.5-7B} \\
        \midrule
        \textcolor{think}{<think>}
        To solve this question, I first need to determine who hosted the 16th Annual Young Hollywood Awards. After identifying the host, I must verify whether the person is a British singer-songwriter.
        \textcolor{think}{</think>} \\

        \textcolor{search}{<search>}
        host of the 16th Annual Young Hollywood Awards
        \textcolor{search}{</search>} \\

        \textcolor{information}{<result>}
        Doc 1 (Title: "Young Hollywood Awards")  
        The Young Hollywood Awards is an annual awards ceremony honoring emerging talent in film, music, and television. The 16th Annual Young Hollywood Awards were held in Los Angeles in 2014 and featured appearances from a variety of young actors and musicians. The ceremony was hosted by Kelly ... Doc 2 (Title: "2014 Young Hollywood Awards")  
        The 2014 ceremony, recognized as the 16th installment of the awards, took place at The Wiltern theater. Television personality and singer Kelly Osbourne served as host for the event. The show highlighted rising stars and ...
        Doc 3 (Title: "Young Hollywood Awards")  
        Previous hosts of the ceremony have included a mixture of actors and entertainers. For the 2014 edition, Kelly Osbourne was selected due to her strong presence in both television and music, making her ...
        \textcolor{information}{</result>} \\

        \textcolor{think}{<think>}
        The retrieved documents consistently identify Kelly Osbourne as the host. Next, I need to confirm whether she is a British singer-songwriter.
        \textcolor{think}{</think>} \\

        \textcolor{search}{<search>}
        Kelly Osbourne nationality singer songwriter
        \textcolor{search}{</search>} \\

        \textcolor{information}{<result>}
        Doc 1 (Title: "Kelly Osbourne")  
        Kelly Lee Osbourne is an English singer, songwriter, television personality, actress, and fashion designer. Born in London, she is the daughter of Ozzy and Sharon Osbourne. She gained prominence through the reality television ...
        Doc 2 (Title: "Kelly Osbourne discography")  
        Osbourne released her debut album *Shut Up* in 2002, which featured pop-rock influences. She later recorded several singles and collaborations, establishing ...
        Doc 3 (Title: "English musicians")  
        England has produced numerous singer-songwriters across genres such as rock, pop, and folk. Kelly Osbourne is frequently categorized among English music artists who ...
        
        \textcolor{information}{</result>} \\

        \textcolor{think}{<think>}
        The evidence confirms that Kelly Lee Osbourne is English (therefore British) and has an established career as a singer-songwriter. Hence, she satisfies all constraints in the question.
        \textcolor{think}{</think>} \\

        \textcolor{answer}{<answer>}
         Kelly Lee Osbourne hosted the 16th Annual Young Hollywood Awards. \boxed{\text{Kelly Lee Osbourne}}
        \textcolor{answer}{</answer>} \\

        \bottomrule
    \end{tabularx}
\end{table*}

\begin{table*}[h]
    \caption{
        Case Study (ii).
    }
    \label{tab:case_study_2}
    \begin{tabularx}{\linewidth}{X}
        \toprule
        \rowcolor{gray!20}
        \textbf{Example from the MuSiQ. dataset} \\ 
        \midrule
        \textbf{Question}: When was the region immediately north of the region prevailing with the disgrace of the Near East and the site of the most growth in desalination for agricultural use established? \\
        \midrule
        \textbf{Ground Truth}: 1932 \\
        \midrule
        \rowcolor{gray!20}
        \textbf{Model Output by RAPO-Qwen2.5-7B} \\
        \midrule

        \textcolor{think}{<think>}
        The question appears to involve multiple geographic references. First, I need to identify the region described as having the “disgrace of the Near East” and being a major site of desalination growth for agriculture. After locating that region, I should determine which region lies immediately north of it, and finally find the establishment date of that northern region.
        \textcolor{think}{</think>} \\

        \textcolor{search}{<search>}
        region disgrace of the Near East desalination agricultural growth
        \textcolor{search}{</search>} \\

        \textcolor{information}{<result>}
        Doc 1 (Title: "Negev")  
        The Negev is a desert region in southern Israel. Historically, parts of the desert were once described as emblematic of the “disgrace of the Near East” due to arid conditions and limited agricultural productivity. In recent decades, however, the Negev has experienced ...
        Doc 2 (Title: "Desalination in Israel")  
        Israel has pioneered desalination technologies to support agriculture in arid regions. Significant growth has occurred in the southern desert areas, particularly the Negev, where desalinated water has ...
        Doc 3 (Title: "Arava Valley")  
        The Arava Valley is another desert area known for agricultural innovation using advanced irrigation techniques, though it is less commonly referenced as the primary site of desalination growth compared ...
        \textcolor{information}{</result>} \\

        \textcolor{think}{<think>}
        The documents consistently indicate that the Negev matches both clues: it was historically characterized negatively and later became a major center for desalination-supported agriculture. Next, I need to identify the region immediately north of the Negev.
        \textcolor{think}{</think>} \\

        \textcolor{search}{<search>}
        region north of the Negev
        \textcolor{search}{</search>} \\

        \textcolor{information}{<result>}
        Doc 1 (Title: "Southern District (Israel)")  
        The Southern District is one of Israel's administrative districts and encompasses most of the Negev desert. Directly north of this district lies the ...
        Doc 2 (Title: "Central District (Israel)")  
        The Central District is an administrative region of Israel located immediately north of the Southern District. It includes major urban centers and ...
        Doc 3 (Title: "Districts of Israel")  
        Israel is divided into several administrative districts, including Northern, Haifa, Central, Tel Aviv, Jerusalem, and Southern. The Central District borders the ...
        \textcolor{information}{</result>} \\

        \textcolor{think}{<think>}
        Since the Southern District contains the Negev and the Central District lies immediately north of it, the target region should be the Central District. Now I need to determine when the Central District was established.
        \textcolor{think}{</think>} \\

        \textcolor{search}{<search>}
        Central District Israel established
        \textcolor{search}{</search>} \\

        \textcolor{information}{<result>}
        Doc 1 (Title: "Central District (Israel)")  
        The Central District is one of Israel’s administrative districts. It was formally established in 1932 during the period of British administration, when regional governance structures ...
        Doc 2 (Title: "Administrative divisions of Mandatory Palestine")  
        During the early 20th century, several administrative regions were created under British rule. Among them was the Central District, established in 1932 as part of a ...
        Doc 3 (Title: "Tel Aviv District")  
        The Tel Aviv District was formed later as administrative needs changed, and therefore does not correspond to the earlier establishment ...
        \textcolor{information}{</result>} \\

        \textcolor{think}{<think>}
        The evidence confirms that the Central District—located immediately north of the Southern District containing the Negev—was established in 1932. Therefore, the answer is 1932.
        \textcolor{think}{</think>} \\

        \textcolor{answer}{<answer>}
        The final answer is \boxed{\text{1932}}.
        \textcolor{answer}{</answer>} \\

        \bottomrule
    \end{tabularx}
\end{table*}

\begin{table*}[]
    \caption{
        Case Study (iii).
    }
    \label{tab:case_study_3}
    \begin{tabularx}{\linewidth}{X}
        \toprule
        \rowcolor{gray!20}
        \textbf{Example from GSM8K dataset} \\ 
        \midrule
        \textbf{Question}: Jan has three times the number of pets as Marcia. Marcia has two more pets than Cindy. If Cindy has four pets, how many total pets do the three have? \\
        \midrule
        \textbf{Ground Truth}: 28 \\
        \midrule
        \rowcolor{gray!20}
        \textbf{Model Output by RAPO-Qwen2.5-7B} \\
        \midrule

        \textcolor{think}{<think>}
        This problem forms a dependency chain: Cindy $\rightarrow$ Marcia $\rightarrow$ Jan. 
        I will compute the intermediate variables first, then aggregate them. 
        \textcolor{think}{</think>} \\

        \textcolor{search}{<python>}
        \texttt{cindy = 4}
        
        \texttt{marcia = cindy + 2}
        
        \texttt{jan = 3 * marcia}
        
        \texttt{print(f"(\{marcia\}, \{jan\})")}
        \textcolor{search}{</python>} \\

        \textcolor{information}{<result>}
        (6, 18)
        \textcolor{information}{</result>} \\

        \textcolor{think}{<think>}
        Now that Cindy = 4, Marcia = 6, and Jan = 18, I compute the total.
        \textcolor{think}{</think>} \\

        \textcolor{search}{<python>}
        \texttt{total = 4 + 6 + 18}
        
        \texttt{print(total)}
        \textcolor{search}{</python>} \\

        \textcolor{information}{<result>}
        28
        \textcolor{information}{</result>} \\

        \textcolor{answer}{<answer>}
        \boxed{\text{28}}
        \textcolor{answer}{</answer>} \\

        \bottomrule
    \end{tabularx}
\end{table*}

\begin{table*}
    \caption{
        Case Study (iv).
    }
    \label{tab:case_study_4}
    \begin{tabularx}{\linewidth}{X}
        \toprule
        \rowcolor{gray!20}
        \textbf{Example from MATH dataset} \\ 
        \midrule
        \textbf{Question}: One computer backs up data every five minutes, while another backs up every two minutes. 
        Find the number of times in twenty-four hours that both computers back up simultaneously. 
        Assume they do not back up at the start of the 24-hour period. \\
        \midrule
        \textbf{Ground Truth}: 144 \\
        \midrule
        \rowcolor{gray!20}
        \textbf{Model Output by RAPO-Qwen2.5-7B} \\
        \midrule

        \textcolor{think}{<think>}
        The computers will back up together at intervals equal to the least common multiple (LCM) of 5 and 2 minutes. 
        I will first compute the LCM to determine the synchronization interval, then calculate how many such intervals occur within 24 hours.
        \textcolor{think}{</think>} \\

        \textcolor{search}{<python>}
        \texttt{import math}
        
        \texttt{interval = math.lcm(5, 2)}
        
        \texttt{print(interval)}
        \textcolor{search}{</python>} \\

        \textcolor{information}{<result>}
        10
        \textcolor{information}{</result>} \\

        \textcolor{think}{<think>}
        The computers synchronize every 10 minutes. 
        Since 24 hours contain $24 \times 60 = 1440$ minutes and the initial moment is excluded, 
        I compute how many 10-minute intervals fit into 1440 minutes.
        \textcolor{think}{</think>} \\

        \textcolor{search}{<python>}
        \texttt{total\_minutes = 24 * 60}
        
        \texttt{total\_minutes // 10}
        
        \texttt{print(total\_minutes)}
        \textcolor{search}{</python>} \\

        \textcolor{information}{<result>}
        144
        \textcolor{information}{</result>} \\

        \textcolor{answer}{<answer>}
        The final answer is \boxed{\text{144}}.
        \textcolor{answer}{</answer>} \\

        \bottomrule
    \end{tabularx}
\end{table*}

\end{document}